\documentclass{article} 
\usepackage{longcat_style,times}

\usepackage{amsmath,amsfonts,bm}

\def\eqref#1{equation~\ref{#1}}

\def\1{\bm{1}}

\DeclareMathAlphabet{\mathsfit}{\encodingdefault}{\sfdefault}{m}{sl}
\SetMathAlphabet{\mathsfit}{bold}{\encodingdefault}{\sfdefault}{bx}{n}

\usepackage{hyperref}
\usepackage{url}

\usepackage{graphicx}
\usepackage{balance}  
\usepackage{amsmath}
\usepackage{algpseudocode}
\usepackage{latexsym}
\usepackage{multirow}
\usepackage{multicol}
\usepackage{color}
\usepackage{dashrule}
\usepackage{extarrows}
\usepackage{dsfont}
\usepackage{centernot}
\usepackage{hyperref}
\usepackage{natbib}
\usepackage{fancybox}
\usepackage[utf8]{inputenc} 
\usepackage[T1]{fontenc}    
\usepackage{hyperref}       
\usepackage{url}            
\usepackage{booktabs}       
\usepackage{amsfonts}       
\usepackage{amsmath}
\usepackage{nicefrac}       
\usepackage{microtype}      
\usepackage{xcolor} 
\usepackage{amsmath}
\usepackage{lipsum}
\usepackage{colortbl}
\usepackage{subcaption}
\usepackage{tabularx}
\usepackage{makecell}
\usepackage{wrapfig}
\usepackage{bm}
\usepackage{multirow}
\usepackage{xcolor} 
\usepackage{bbm}
\usepackage{mdframed}
\usepackage{enumitem}
\usepackage{xspace}

\usepackage{lineno}

\definecolor{darkblue}{rgb}{0, 0, 0.5}
\hypersetup{colorlinks=true, citecolor=darkblue, linkcolor=darkblue, urlcolor=darkblue}

\usepackage{listings}
\usepackage{xcolor}
\usepackage[ruled,vlined]{algorithm2e}
\usepackage{caption}
\usepackage{float} 
\usepackage{arydshln}
\usepackage{wrapfig} 
\usepackage{tcolorbox}
\tcbuselibrary{theorems}
\tcbuselibrary{most}
\usepackage[dvipsnames]{xcolor}

\usepackage{hyperref}       
\usepackage{url}            
\usepackage{booktabs}       
\usepackage{amsfonts}       
\usepackage{nicefrac}       
\usepackage{microtype}      
\usepackage{xcolor}         
\usepackage{amsmath}
\usepackage{multirow}
\usepackage{multicol}
\usepackage{colortbl}
\usepackage{tcolorbox}
\usepackage{wrapfig}

\usepackage{cleveref}
\usepackage{xspace}

\makeatletter
\DeclareRobustCommand\onedot{\futurelet\@let@token\@onedot}
\def\@onedot{\ifx\@let@token.\else.\null\fi\xspace}

\def\eg{\emph{e.g}\onedot}

\def\etc{\emph{etc}\onedot}

\makeatother

\usepackage{natbib}

\definecolor{light-gray}{gray}{0.6}
\definecolor{front-color}{HTML}{F5FFFA}
\definecolor{Gray}{gray}{0.93}

\definecolor{customTeal}{RGB}{0, 128, 128} 
\definecolor{emphasisColor}{RGB}{255, 0, 0} 
\definecolor{oursBlue}{RGB}{51,202,246}

\definecolor{blue1}{HTML}{508AB2}
\definecolor{green2}{HTML}{BFF6BA}

\newtcbtheorem[]{prompt}{Prompt}%
{colback=SeaGreen!10!CornflowerBlue!10,
 colframe=RoyalPurple!55!Aquamarine!100!,
 fonttitle=\bfseries,
 left=.02in, right=.02in, bottom=.02in, top=.02in,
 before upper={\linespread{1.5}\selectfont}}{prompt}

\renewcommand{\shorttitle}{OneThinker: All-in-one Reasoning Model for Image and Video}

\definecolor{darkblue}{rgb}{0, 0, 0.5}
\hypersetup{colorlinks=true, citecolor=darkblue, linkcolor=darkblue, urlcolor=darkblue}

\makeatletter
\renewcommand{\@maketitle}{%
  \vbox{%
    \hsize\textwidth
    \linewidth\hsize
    \vskip -0.5in
    \noindent
    \begin{minipage}{0.99\textwidth}
  \includegraphics[width=0.27\linewidth]{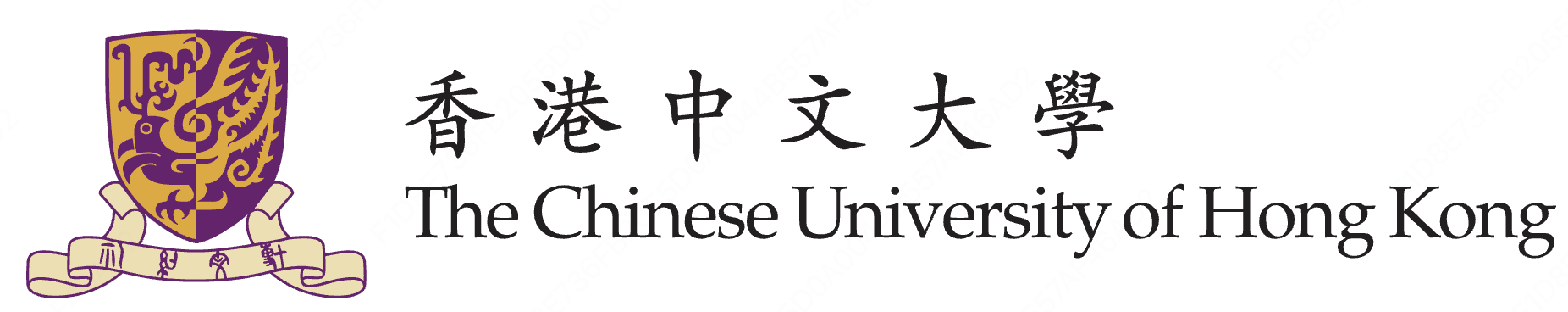}
    \end{minipage}%
    \\
    \rule{\linewidth}{1pt}
    \hspace{0.05\textwidth}%
    \begin{minipage}{0.8\textwidth}
    \end{minipage}

    \centering
    {\LARGE \bfseries\@title\par}
    \vskip 0.1in  
    \def\And{%
      \end{tabular}\hfil\linebreak[0]\hfil%
      \begin{tabular}[t]{c}\bf\rule{\z@}{24\p@}\ignorespaces%
    }
    \def\AND{%
      \end{tabular}\hfil\linebreak[4]\hfil%
      \begin{tabular}[t]{c}\bf\rule{\z@}{24\p@}\ignorespaces%
    }
    \begin{tabular}[t]{c}\bf\rule{\z@}{24\p@}\@author\end{tabular}%
  \vskip 0.05in 
  }
}
\makeatother

\title{OneThinker: All-in-one Reasoning Model for Image and Video \\}

\makeatletter
\def\@fnsymbol#1{\ensuremath{\ifcase#1\or \dagger\or \ddagger\or
   \mathsection\or \mathparagraph\or \|\or **\or \dagger\dagger
   \or \ddagger\ddagger \else\@ctrerr\fi}}
\makeatother

\newcommand{\homepage}{\raisebox{-1.5pt}{\includegraphics[height=1em]{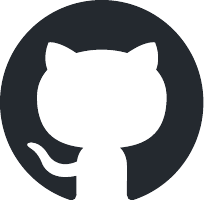}}}
\newcommand{\hfmodel}{\raisebox{-1.5pt}{\includegraphics[height=1em]{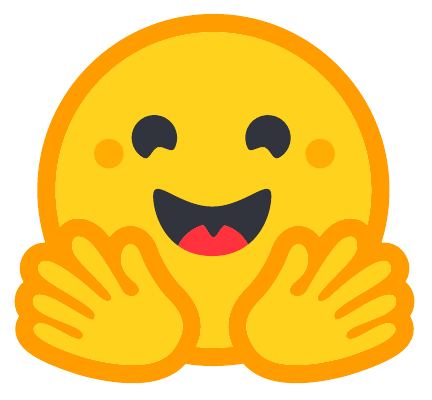}}}

\author{
\begin{tabular}{c}
\textbf{Kaituo Feng}$^{1,2}$
\quad
\textbf{Manyuan Zhang}$^{2}$\thanks{Project Leader.}
\quad
\textbf{Hongyu Li}$^{2}$
\quad
\textbf{Kaixuan Fan}$^{1,2}$
\quad
\textbf{Shuang Chen}$^{2}$ \\[1ex]
\textbf{Yilei Jiang}$^{1,2}$
\quad
\textbf{Dian Zheng}$^{1,2}$
\quad
\textbf{Peiwen Sun}$^{1}$
\quad
\textbf{Yiyuan Zhang}$^{1}$
\quad
\textbf{Haoze Sun}$^{2}$ \\[1ex]
\textbf{Yan Feng}$^{2}$ 
\quad
\textbf{Peng Pei}$^{2}$
\quad
\textbf{Xunliang Cai}$^{2}$
\quad
\textbf{Xiangyu Yue}$^{1}$\thanks{Corresponding Author.} \\[1ex]
\normalfont $^1$MMLab, CUHK 
\quad
$^2$Meituan\\[1ex]
{\homepage\ \normalfont 
\texttt{Home: \!\!\!\!\!\url{https://github.com/tulerfeng/OneThinker}}} \\
{\hfmodel\ \normalfont \texttt{HF: \!\!\!\url{https://huggingface.co/OneThink}}} \\
\end{tabular}
}

\begin{document}

{%
   \renewcommand\twocolumn[1][]{#1}%
   \maketitle
   \vspace{-1pt}
   \begin{center}
    \centering
    \includegraphics[width=0.99\linewidth]{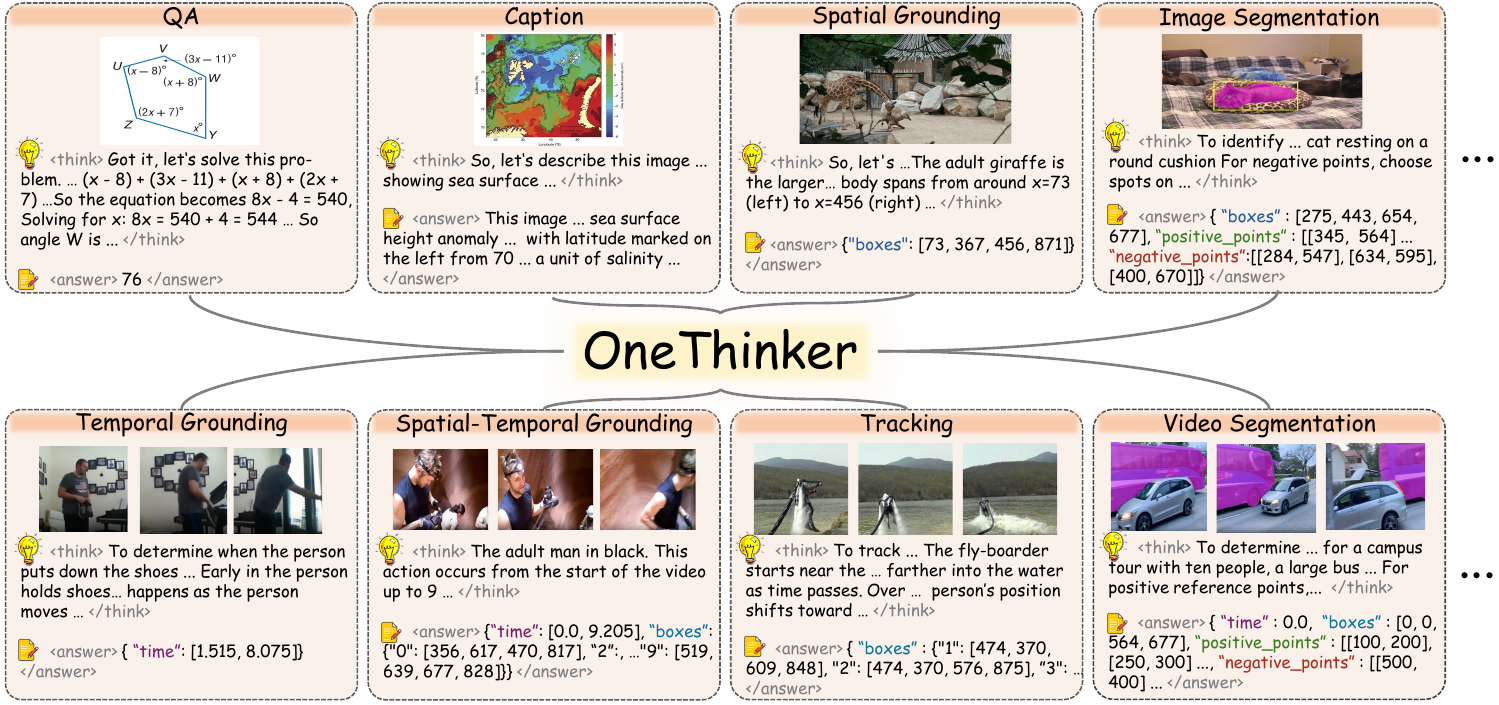
    }
    \captionof{figure}{Overview of our OneThinker, which is capable of thinking across a wide range of  tasks for image and video understanding.}
    \vspace{6pt}
    \label{fig:teaser}
   \end{center}%
  }

\begin{abstract}
Reinforcement learning (RL) has recently achieved remarkable success in eliciting visual reasoning within Multimodal Large Language Models (MLLMs). However, existing approaches typically train separate models for different tasks and treat image and video reasoning as disjoint domains. This results in limited scalability toward a multimodal reasoning generalist, which restricts practical versatility and hinders potential knowledge sharing across tasks and modalities.
To this end, we propose OneThinker, an all-in-one reasoning model that unifies image and video understanding across diverse fundamental visual tasks, including question answering, captioning, spatial and temporal grounding, tracking, and segmentation. To achieve this, we construct the OneThinker-600k training corpus covering all these tasks and employ commercial models for CoT annotation, resulting in OneThinker-SFT-340k for SFT cold start.
Furthermore, we propose EMA-GRPO to handle reward heterogeneity in multi-task RL by tracking task-wise moving averages of reward standard deviations for balanced optimization. Extensive experiments on diverse visual benchmarks show that OneThinker delivers strong performance on 31 benchmarks, across 10 fundamental visual understanding tasks.
Moreover, it exhibits effective knowledge transfer between certain tasks and preliminary zero-shot generalization ability, marking a step toward a unified multimodal reasoning generalist. All code, model, and data are released.

\end{abstract}    
\section{Introduction}
\label{sec:intro}

\begin{figure*}
  \centering
  \includegraphics[width=0.99\linewidth]{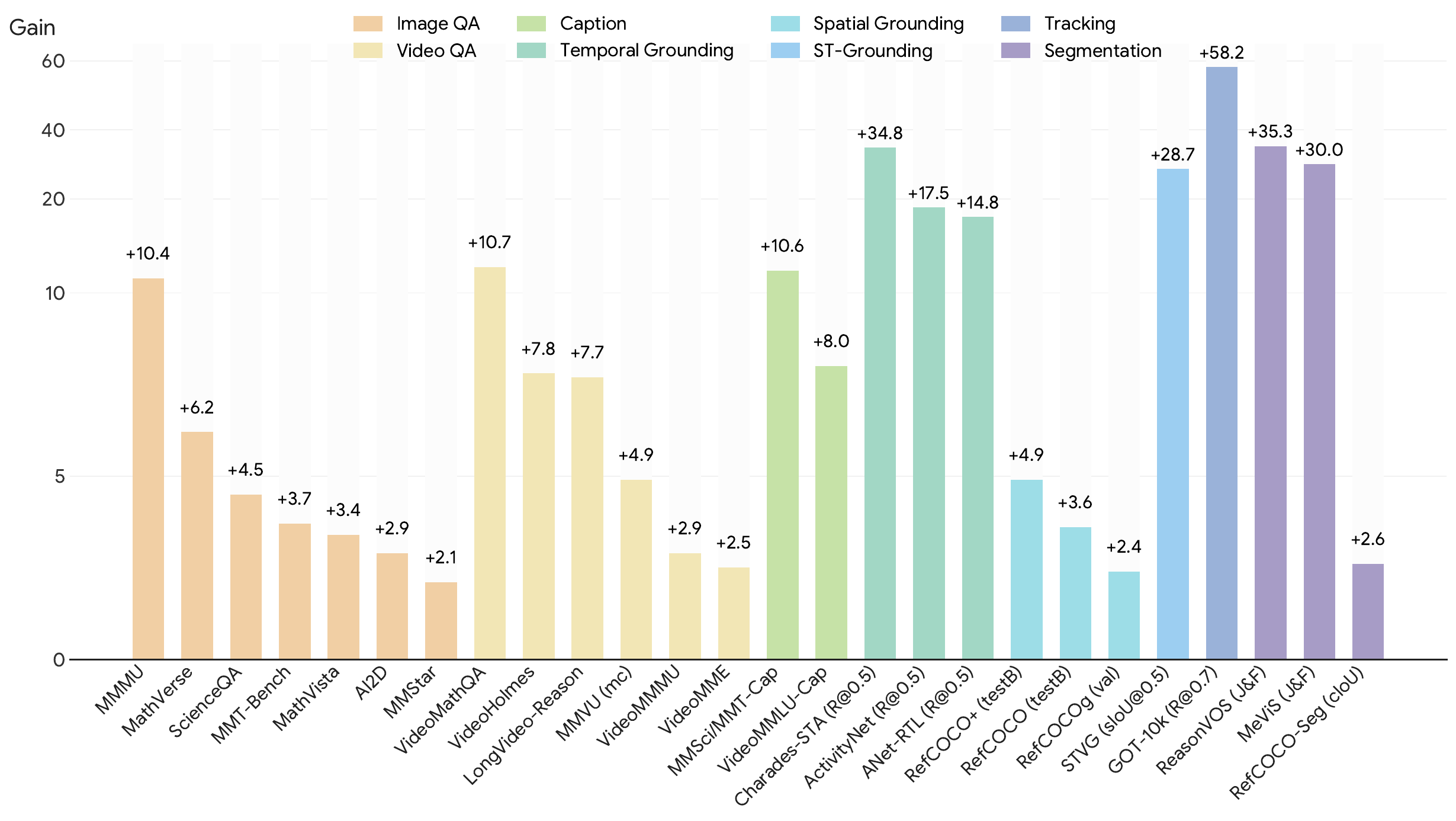}
  \caption{
Performance gains of our model over Qwen3-VL-Instruct-8B across diverse visual tasks after training.
  }
  \vspace{-0.15in}
\end{figure*}

Reasoning serves as a cornerstone in advancing Multimodal Large Language Models (MLLMs) toward artificial general intelligence (AGI), enabling them to perform step-by-step inference over complex visual–linguistic inputs \cite{zhang2023multimodal,yuan2025mme,zhou2025reinforced}.  Inspired by DeepSeek-R1 \cite{guo2025deepseek}, a growing number of studies have witnessed the success of adopting reinforcement learning (RL) with the Group Relative Policy Optimization (GRPO) algorithm to enhance reasoning abilities \cite{wu2025reinforcing,liu2025seg,zhang2025critique,li2025star,feng2025video}. For instance, Vision-R1 \cite{huang2025vision} and Video-R1 \cite{feng2025video} demonstrate strong reasoning performance on image and video question answering, respectively, while VLM-R1 \cite{shen2025vlm} excels in image detection and Seg-R1 \cite{you2025seg} in segmentation. These advances underscore the remarkable effectiveness and broad potential of RL-based training for a wide range of visual tasks.

However, existing thinking models are usually designed to handle only a single task and operate exclusively on either images or videos. 
Such separation greatly limits their practical versatility and may also hinder the potential benefits of cross-task and cross-modal knowledge transfer.
Although a few works have explored extending MLLMs with RL for multiple tasks \cite{li2025videochat,yu2025perception,zhang2025thinking}, they are usually confined to limited subsets of visual tasks within a single modality. Furthermore, these approaches are often constrained by small-scale tuning, which limits their ability to generalize beyond specific domains. For instance, VideoChat-R1 \cite{li2025videochat} performs co-training on only three spatio-temporal perception tasks with merely 18k samples, and remains restricted to the video modality.
Recognizing that vision inherently encompasses both static images and dynamic videos, and that real-world scenarios demand unified reasoning across diverse visual tasks, we pose a question: 

\textit{Can we train an all-in-one multimodal reasoning generalist, which is capable of simultaneously handling both image and video understanding across diverse fundamental visual tasks?}

To achieve this, we present OneThinker, a unified multimodal reasoning generalist capable of handling a wide range of visual reasoning tasks, including question answering, captioning, spatial and temporal grounding, tracking, and segmentation. 
First, we curate a large-scale dataset OneThinker-600k, comprising approximately 600k multimodal samples that jointly cover these fundamental visual tasks. We then employ a strong proprietary model Seed1.5-VL \cite{guo2025seed1} to annotate and filter high-quality chain-of-thought (CoT) data, resulting in OneThinker-SFT-340k dataset for SFT cold start. Through joint multi-task training across both images and videos, OneThinker effectively learns to reason over spatial and temporal cues in a unified manner.

Besides, considering the distinct reward characteristics of heterogeneous visual tasks, we further introduce EMA-GRPO to improve RL training. This is motivated by two complementary imbalances: Standard GRPO suffers from intra-task imbalance because its sample-wise standard deviation (std) normalization favors low-variance rollouts \cite{liu2025understanding,bereket2025uncalibrated, chu2025gpg,huang2025mapo}; Conversely, removing this std normalization, as in Dr.GRPO \cite{liu2025understanding}, causes inter-task imbalance, where sparse-reward tasks (\eg, math)  dominate while dense ones (\eg, detection) are suppressed.
EMA-GRPO addresses both issues by maintaining task-wise exponential moving averages of reward standard deviations for normalization.
This design allows each task to have a stable yet adaptive normalization scale that reflects its own reward dynamics.
It balances intra-task weighting by reducing bias toward low-variance samples and prevents inter-task imbalance by using independent normalization statistics for different tasks, resulting in stable and balanced optimization across diverse visual tasks.

Extensive experiments demonstrate that OneThinker achieves consistently strong performance across diverse visual reasoning benchmarks. For example, OneThinker-8B reaches 70.6\% accuracy on MMMU \cite{yue2024mmmu} and 64.3\% on MathVerse \cite{zhang2024mathverse} for image QA.
In perception-oriented tasks such as grounding, tracking, and segmentation, our model also delivers
strong results, for example, 84.4 R@0.5 on GOT-10k \cite{huang2019got} and 54.9 J\&F on ReasonVOS \cite{bai2024one}.
Moreover, unified training across tasks and modalities encourages effective knowledge sharing, allowing the model to transfer reasoning skills between several related tasks and exhibit preliminary zero-shot generalization on unseen scenarios.

Our main contributions are summarized as follows:

\begin{itemize}
    \item We propose OneThinker, a unified multimodal reasoning generalist that handles a wide range of image and video tasks within a single model, including question answering, captioning, grounding, tracking, and segmentation. To support training, we construct the large-scale datasets OneThinker-600k and its CoT-annotated subset OneThinker-SFT-340k.
    
    \item To address the distinct reward characteristics of heterogeneous visual tasks, we introduce EMA-GRPO, which mitigates both intra-task and inter-task imbalance through task-wise adaptive normalization of reward statistics.
    
    \item Extensive experiments demonstrate that OneThinker achieves superior results on 31 benchmarks, across 10 fundamental visual understanding tasks. Besides, it promotes effective  knowledge sharing in certain tasks, and exhibits preliminary zero-shot generalization abilities.
\end{itemize}

\section{Related Works}
\label{sec:related}

\subsection{Reinforcement Learning for LLM Reasoning}

Reinforcement learning (RL) has emerged as a powerful technique for enhancing the reasoning capabilities of Large Language Models (LLMs) \cite{zhang2025critique,zheng2025group,dong2025agentic,yu2025dapo,xie2025logic}.
Recent studies, exemplified by DeepSeek-R1, adopt rule-based RL with Group Relative Policy Optimization (GRPO) \cite{guo2025deepseek}  algorithm to directly optimize outcome-level rewards, enabling step-by-step reasoning without explicit intermediate supervision. The success of DeepSeek-R1 motivates a surge of works exploring this paradigm further \cite{zhang2025critique,chen2025towards,feng2025group}. For example, Dr.GRPO \cite{liu2025understanding} introduces an unbiased optimization method that addresses the sample-wise standard deviation imbalance and response-length bias inherent in standard GRPO. Besides, GSPO \cite{zheng2025group} introduces a sequence-level RL algorithm that replaces token-wise ratios with sequence-level optimization, improving training stability for large-scale Mixture-of-Experts models. Critique-GRPO \cite{zhang2025critique} integrates natural language feedback to guide policy optimization, enabling LLMs to refine their reasoning through critique-based self-improvement beyond standard RL fine-tuning.
However, most existing research still focuses on single tasks or homogeneous reasoning objectives, where reward distributions remain relatively consistent.

\subsection{Reasoning in MLLMs}

Inspired by the success of reasoning in LLMs, a rising trend of works aims to bring this capability into MLLMs, enabling reasoning in different visual tasks \cite{li2025star,sun2025reinforcement,feng2025video,sun2025spacevista,zhou2025reinforced,duan2025codeplot,chen2025advancing,chen2025ares,meng2025open}. 
For instance, Vision-R1 \cite{huang2025vision} tackles complex image reasoning in visual question answering, while Video-R1 \cite{feng2025video} advances question answering over dynamic video inputs. 
Perception-R1 \cite{yu2025perception} and VLM-R1 \cite{shen2025vlm} further extend this paradigm to image object detection, revealing the potential of RL for perception-oriented tasks.
Seg-R1 \cite{you2025seg} introduces a decoupled reasoning–segmentation framework that employs GRPO-based RL to generate explicit chain-of-thought reasoning and positional prompts for image segmentation tasks.
Time-R1 \cite{wang2025time} adapts RL-based post-training to temporal grounding in videos and achieves promising results, whereas VideoChat-R1 \cite{li2025videochat} applies reinforcement fine-tuning on three spatio-temporal tasks to enhance perception and reasoning in video understanding.  SophiaVL-R1 \cite{fan2025sophiavl} introduces thinking-process rewards to improve RL training for image question answering. While these approaches have achieved remarkable progress in multimodal reasoning, most models remain restricted to limited tasks, and support either image or video reasoning alone.
\section{Method}
\label{sec:method}

\subsection{Dataset Construction}

\paragraph{Data Collection and Curation.}  
High-quality and diverse training data are essential for developing a unified multimodal reasoning generalist. To this end, we construct the \textbf{OneThinker-600k} corpus as the foundation for training, as illustrated in \cref{dataset_info}.
Our dataset covers both image and video modalities and spans a series of fundamental visual reasoning tasks, including rule-based QA, open-ended QA, captioning, spatial grounding, temporal grounding, spatio-temporal grounding, tracking, and segmentation. 
For perception-oriented tasks such as grounding, tracking, and segmentation, we require the model to output responses in a predefined JSON schema to ensure consistent formatting and enable automatic, verifiable reward computation. 
Details of prompts and formats are provided in the Appendix.  

To ensure task diversity and balanced modality coverage, we collect data from a broad range of public training datasets and carefully curate samples across various domains and difficulty levels. 
The curated dataset is designed to equip the model with a broad spectrum of core reasoning abilities, such as logical reasoning, knowledge-based inference, spatial perception, temporal understanding, causal inference, \etc Together, these capabilities enable a unified multimodal reasoning generalist that can perform structured and coherent inference over both static and dynamic visual contexts.

\paragraph{CoT Annotation.}  To enable effective SFT initialization for reasoning, we leverage a strong proprietary model, \textbf{Seed1.5-VL} \cite{guo2025seed1}, to produce CoT annotations on the previously constructed OneThinker-600k corpus.
For different tasks, we apply task-specific filtering thresholds to ensure the accuracy of retained CoT traces.
After rule-based checking and quality validation, we obtain the CoT-annotated subset \textbf{OneThinker-SFT-340k}. This SFT dataset provides a diverse and reliable foundation for developing unified multimodal reasoning across a wide range of visual tasks.

\begin{figure*}
  \centering
  \includegraphics[width=0.99\linewidth]{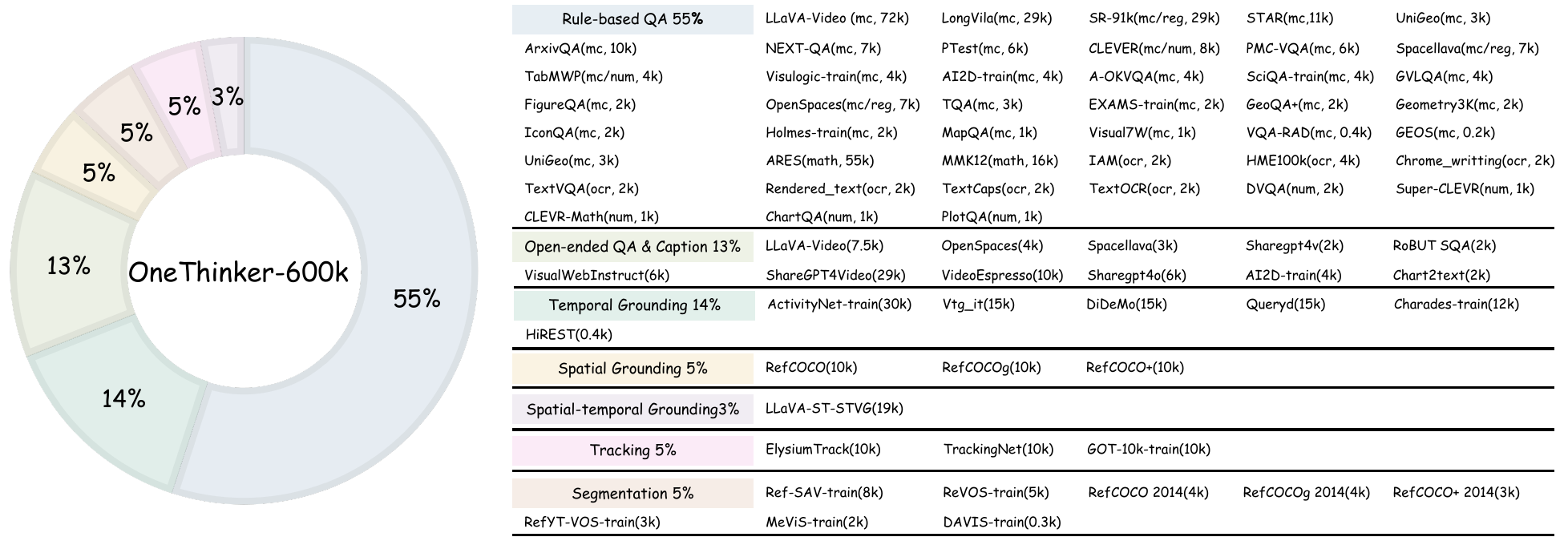}
  \caption{
Overview of our curated training dataset, including both image and video modalities for a diverse range of understanding tasks.
  }
  \label{dataset_info}
  \vspace{-0.15in}
\end{figure*}

\subsection{Task Types and Rewards}

All tasks are cast into a unified text interface, where the model first produces its internal reasoning
inside \texttt{<think>...</think>} and then outputs a task-specific result inside
\texttt{<answer>...</answer>}. For perception-oriented tasks, the \texttt{<answer>} block contains
a structured representation (\eg, time spans, bounding boxes, sparse points) following a predefined
schema, which allows automatic parsing and verification. 
The overall reward is
\begin{equation}
  R = R_{\text{acc}} \;+\; R_{\text{format}}\,,
\end{equation}
where $R_{\text{acc}}$ is task-specific accuracy reward and $R_{\text{format}}$ is format reward.  For tasks requiring structured outputs, $R_{\text{format}}$ further checks whether the output follows the predefined schema.

\paragraph{Rule-based QA.}
This category includes multiple-choice, numerical, regression, math, and OCR tasks.
For multiple-choice, numerical, and math problems, correctness is determined by whether the predicted
and ground-truth answers are equivalent. Regression tasks are evaluated using the Mean Relative
Accuracy (MRA) metric \cite{yang2025thinking}, which measures relative closeness between the prediction and the reference
value across multiple tolerance levels. OCR tasks use the Word Error Rate to compute the reward.  
These rule-based tasks provide deterministic and interpretable feedback for discrete reasoning and
quantitative prediction, forming a reliable foundation for reinforcement learning.

\paragraph{Open-ended QA \& Caption.}
For open-ended question answering and captioning tasks, we employ an external reward model to provide a similarity score:
\begin{equation}
  R_{\text{acc}} \;=\; \mathrm{RM}\bigl(q,\, \hat{a},\, a\bigr),
\end{equation}
where $q$ denotes the input query, $\hat{a}$ is the model-predicted answer, and $a$ is the reference answer.  
In this work, we adopt POLAR-7B \cite{dou2025pre} as the reward model $\mathrm{RM}$.

\paragraph{Temporal Grounding.}
Temporal grounding requires the model to identify the start and end time of the queried event in a
video. The answer encodes a continuous time segment, and we measure accuracy using temporal IoU:
\begin{equation}
  R_{\text{acc}} \;=\; \mathrm{tIoU}\bigl([\hat{s}, \hat{e}], [s, e]\bigr),
\end{equation}
where $\mathrm{tIoU}(\cdot,\cdot)$ denotes the temporal intersection-over-union of two intervals.
Here, $\hat{s}$ and $\hat{e}$ represent the predicted start and end timestamps, while $s$ and $e$
denote their corresponding ground-truth values.

\paragraph{Spatial Grounding.}
Spatial grounding requires the model to localize a target region by predicting a bounding box. The accuracy is measured using spatial intersection-over-union (sIoU)
between predicted and ground-truth boxes:
\begin{equation}
  R_{\text{acc}} \;=\; \mathrm{sIoU}\bigl(\hat{b}, b\bigr),
\end{equation}
where $\hat{b}$ and $b$ denote the predicted and ground-truth bounding boxes, respectively, and
$\mathrm{sIoU}(\cdot,\cdot)$ represents their spatial overlap ratio.

\paragraph{Spatial-temporal Grounding.} 
This task unifies temporal and spatial localization, requiring the model to predict
both the temporal span of an event and the corresponding bounding boxes across frames. The accuracy
is computed by combining temporal IoU and mean spatial IoU:
\begin{equation}
  R_{\text{acc}} \;=\;
  \mathrm{tIoU}\bigl([\hat{s}, \hat{e}], [s, e]\bigr)
  \;+\;
  \overline{\mathrm{sIoU}},
\end{equation}
where $\hat{s}$ and $\hat{e}$ denote the predicted start and end times, and $\overline{\mathrm{sIoU}}$
represents the mean IoU between predicted and ground-truth boxes across frames.

\paragraph{Tracking.}
Tracking requires the model to predict a sequence of bounding boxes for a given target across
video frames. The accuracy is measured as the mean IoU over all frames:
\begin{equation}
  R_{\text{acc}} \;=\; \overline{\mathrm{sIoU}},
\end{equation}
where $\overline{\mathrm{sIoU}}$ is the averaged IoU between predicted and ground-truth bounding
boxes throughout the trajectory.

\paragraph{Segmentation.}
Following prior works applying RL for image segmentation \cite{you2025seg,liu2025seg,wang2025affordance}, the model predicts a bounding box along with a set of positive and negative points to identify target objects.
These predictions are subsequently fed into SAM2 \cite{ravi2024sam} to generate the final segmentation mask.
For video segmentation, we further require the model to predict a keyframe time $\hat{t}$ indicating when the predicted boxes and points should be applied.
Due to the high computational latency of running SAM2 on all rollouts for video segmentation, we omit the mask-based reward in this paper. All bounding boxes and point annotations are provided by Seed1.5-VL \cite{guo2025seed1}.

We define a Gaussian kernel
$\mathcal{G}(d) = \exp\!\left(-\tfrac{d^2}{2\sigma^2}\right)$
that normalizes distances into $[0,1]$. We set $\sigma = 50$ for spatial distances and $\sigma = 1$ for temporal distances.

For image segmentation, the accuracy reward combines bounding box overlap with Gaussian similarities
over positive and negative point sets:
\begin{equation}
  R_{\text{acc}}
  \;=\;
  \ \,\mathrm{sIoU}(\hat{b}, b)
  \;+\;
 \,\mathcal{G}(\mathrm{dis}_{+})
  \;+\;
\,\mathcal{G}(\mathrm{dis}_{-}),
\end{equation}
where $\mathrm{dis}_{+}$ denotes the minimum average distance between predicted and ground-truth positive points under optimal matching, and $\mathrm{dis}_{-}$ is defined similarly for negative points.

For video segmentation, a temporal Gaussian kernel is additionally applied to the predicted keyframe
time:
\begin{equation}
  R_{\text{acc}}
  \!=\!\
  \,\mathrm{sIoU}(\hat{b}, b)
  +
 \,\mathcal{G}(|\hat{t}-t|)
  +
\,\mathcal{G}(\mathrm{dis}_{+})
  +
\,\mathcal{G}(\mathrm{dis}_{-}),
\end{equation}
where $\hat{t}$ denotes the predicted keyframe timestamp and $t$ is the annotated ground-truth time. In this paper, the number of positive points and negative points are both set to three. 

\begin{figure*}
  \centering
  \includegraphics[width=0.99\linewidth]{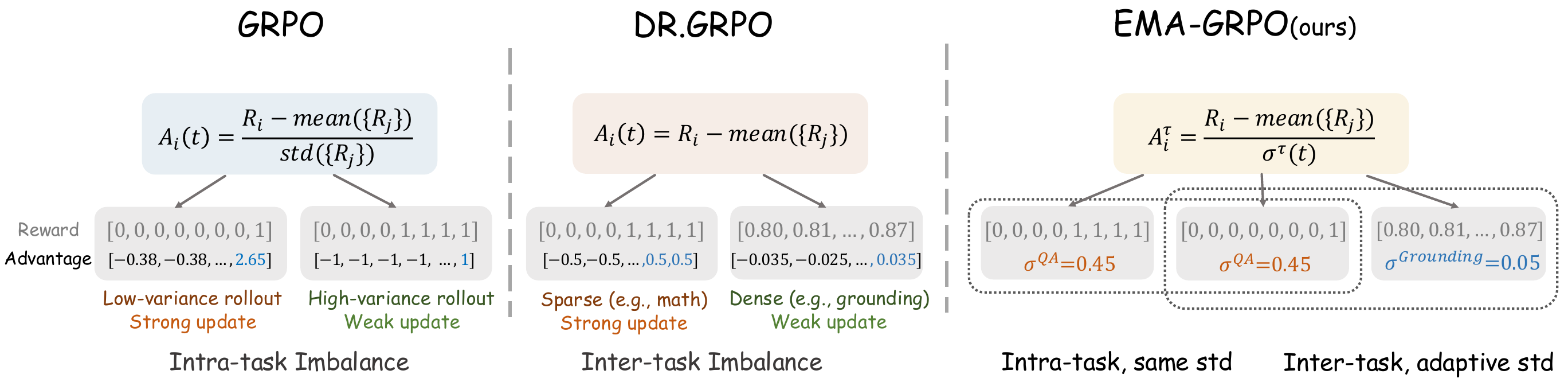}
  \caption{
    Comparison of advantage formulations in three RL algorithms.
  }
  \vspace{-0.1in}
    \label{rlcmp}
\end{figure*}

\subsection{EMA-GRPO}

While GRPO has demonstrated strong capability in enhancing reasoning performance,
its direct application to heterogeneous visual tasks would lead to biased optimization.
We identify two complementary sources of imbalance that hinder effective multi-task training, as illustrated in \cref{rlcmp}.

\paragraph{Intra-task Imbalance.}
Standard GRPO normalizes rewards within each prompt group by the group standard deviation to stabilize optimization.
This normalization causes biased weighting among samples of the same task \cite{liu2025understanding,bereket2025uncalibrated, chu2025gpg,huang2025mapo}.
Specifically, examples with very small or very large variance receive stronger updates, while medium-difficulty samples—whose rollouts usually have large variance—are under-optimized.
As a result, the reinforcement learning within a task becomes biased.

\paragraph{Inter-task Imbalance.}
Conversely, removing the STD normalization as in Dr.GRPO \cite{liu2025understanding} avoids the intra-task bias but introduces
cross-task imbalance: different tasks vary in their reward scale and density, so sparse rewards (\eg, math reasoning) dominate the optimization signal, whereas dense, small-range rewards (\eg, grounding) are down-weighted.
This imbalance causes the model to overfit a small subset of tasks and weakens its generalization
across diverse visual reasoning settings.


\paragraph{EMA-based Normalization.}
To overcome both imbalances, we propose EMA-GRPO, which introduces task-wise adaptive normalization based on the exponential moving average (EMA) of reward statistics.
For each task $\tau$, we maintain EMA estimates of the first and second moments of its outcome rewards.
Given the current batch of rewards $\{R_i\}$ belonging to task $\tau$, let first-order moment
$\mu^{\tau}(t) = \mathrm{mean}(\{R_i\})$ and second-order moment
$\nu^{\tau}(t) = \mathrm{mean}(\{R_i^2\})$ at step $t$.
We update the EMA moments as
\begin{equation}
\begin{aligned}
m_1^{\tau}(t) &= \beta \, m_1^{\tau}(t-1) + (1-\beta)\, \mu^{\tau}(t),\\
m_2^{\tau}(t) &= \beta \, m_2^{\tau}(t-1) + (1-\beta)\, \nu^{\tau}(t),
\end{aligned}
\end{equation}
where $\beta$ is the decay factor (set to $0.99$).
The task-wise standard deviation is then computed as
\begin{equation}
\sigma^{\tau}(t)
= \sqrt{(m_2^{\tau}(t) - (m_1^{\tau}(t))^2}.
\end{equation}
This moving statistic captures each task's intrinsic reward scale while adapting smoothly to changing reward distributions during training.

Then, the advantage in task $\tau$ is computed with its task-wise EMA standard deviation:
\begin{equation}
A_i^{\tau}(t) = \frac{R_i - \mathrm{mean}(\{R_j\})}{\sigma^{\tau}(t)}.
\end{equation}
This adaptive normalization simultaneously resolves both intra-task and inter-task imbalance.
Within each task, all rollouts share the same normalization scale $\sigma^{\tau}(t)$, which prevents
the model from overemphasizing easy or hard samples while under-optimizing medium-difficulty ones.
Across different tasks, each task maintains its own reward scale through an independent
$\sigma^{\tau}(t)$, ensuring balanced gradient contributions regardless of differences in reward
magnitude or density. For numerical stability during the initial stage, when $\sigma^{\tau}(t)$ has not yet stabilized, we clip the advantage to $[-5, 5]$.
Together, these properties promote stable optimization and fair learning across heterogeneous
visual reasoning tasks.

\paragraph{Training Objective.}
Following DeepSeek-R1, the final policy update adopts the standard GRPO objective with the EMA-normalized advantage:
\begin{equation}
\begin{aligned}
&\mathbb{E}_{q,\{o_i\}}\!\Bigg[
\frac{1}{G}\sum_{i=1}^{G}
\Big(
\min\!\Big(
\tfrac{\pi_{\theta}(o_i|q)}{\pi_{\theta_{\text{old}}}(o_i|q)} A_i^{\tau}(t),\; \\[-2pt]
&\qquad\qquad
\mathrm{clip}\!\Big(
\tfrac{\pi_{\theta}(o_i|q)}{\pi_{\theta_{\text{old}}}(o_i|q)},
1-\epsilon,\,1+\epsilon
\Big) A_i^{\tau}(t)
\Big)\\
&\qquad \qquad
-\beta_{\mathrm{KL}}\,
D_{\mathrm{KL}}\!\big(\pi_{\theta}\,\Vert\,\pi_{\mathrm{ref}}\big)
\Big)
\Bigg].
\end{aligned}
\end{equation}
The definition of variables and hyperparameters follows the standard GRPO \cite{guo2025deepseek}.

\section{Experiments}
\label{sec:experiment}

\subsection{Setup}

\textbf{Training Details.} Our model is trained on 32 NVIDIA H800 GPUs. 
In the SFT stage, we adopt Qwen-3-VL-Instruct-8B \cite{qwen3vl} as the base model and train it on our OneThinker-SFT-340k dataset. 
Subsequently, reinforcement learning is performed based on the SFT-initialized model using the OneThinker-600k corpus. 
For both SFT and RL, we sample image-video balanced sets for training.
The batch size is set to 32 for SFT and 128 for RL. 
The learning rate is configured as $1\times10^{-5}$ for SFT and $2\times10^{-6}$ for RL, both optimized with AdamW. For efficiency, the maximum number of video frames is capped at 128 during training. 
The decay factor $\beta$ is set to $0.99$, following the common practice in EMA.
The group size for EMA-GRPO is set to 8, and the KL regularization coefficient $\beta_{\mathrm{KL}}$ is fixed at 0.01. 
The maximum response length is limited to 4096 tokens. 
We discard rollouts that are entirely correct or incorrect during RL training, following the practice in \cite{yu2025dapo}.
Overall, the complete training process takes approximately 10 days.

\textbf{Benchmarks.}
For evaluation, we adopt a variety of benchmarks corresponding to different visual reasoning tasks,
covering question answering, captioning, spatial and temporal grounding, tracking, and segmentation, as presented in the experimental tables.
For Qwen3-VL-Instruct, we report our reproduced results.
We evaluate models using greedy decoding, following prior works \cite{wang2025vl,feng2025video,xiao2025proxythinker}.

\subsection{Main Results}

We evaluate OneThinker across a wide range of visual reasoning benchmarks covering both image and
video modalities, as summarized in \cref{imageqa}, \cref{videoqa}, \cref{caption}, \cref{temporal}, \cref{spatial}, \cref{spatial-temporal}, \cref{tracking}, and \cref{segmentation}.  
Across all benchmarks, OneThinker demonstrates substantial improvements, showcasing its unified and transferable reasoning ability across tasks and modalities. Examples of reasoning responses for each task can be found in Appendix.

\begin{table*}[]
\caption{Performance of different models on image question answering benchmarks. For Qwen3-VL-Instruct-8B, we report our reproduced results under the same setting.}
\label{imageqa}
    \resizebox{\linewidth}{!}{%
    \setlength{\tabcolsep}{0.5mm}
     \renewcommand\arraystretch{1.2}
     \tiny
\begin{tabular}{@{}ccccccccc@{}}
\toprule
\multirow{2}{*}{\textbf{Models}} & \multicolumn{8}{c}{\textbf{Image QA}}                                          \\ \cmidrule(l){2-9} 
                                 & MMMU \cite{yue2024mmmu} & MathVista \cite{lu2023mathvista} & MathVerse \cite{zhang2024mathverse} & MMBench \cite{liu2024mmbench}  & MMStar \cite{chen2024we} & ScienceQA \cite{lu2022learn} & AI2D \cite{kembhavi2016diagram} & MMT-Bench \cite{ying2024mmt} \\ \midrule
\rowcolor{Gray} GPT-4o \cite{hurst2024gpt}                          & 70.7 & 63.8      & 41.2      & 84.3    & 65.1   & 90.1      & 84.9 & 67.7      \\
\rowcolor{Gray} Gemini 2.5 Pro \cite{comanici2025gemini}                  & 81.7 & 82.7      & -         & 90.1    & 77.5   & -         & 88.4 & -         \\
\rowcolor{Gray} Seed1.5-VL \cite{guo2025seed1}                      & 77.9 & 85.6      & -         & 89.9    & 77.8   & -         & 87.3 &           \\ \midrule
SophiaVL-R1-7B \cite{fan2025sophiavl}                   & 61.3 & 71.3      & 48.8      & 85.4    & 66.7   & 90.9      &      & 62.7      \\
Vision-R1-7B \cite{huang2025vision}                    & -    & 73.5      & 52.4      & -       & -      & -         & -    & -         \\
MM-Eureka-7B \cite{meng2025mm}                     & 57.3 & 73.0      & 50.3      & -       & 64.4   &           & -    & -         \\
VL-Rethinker-7B \cite{wang2025vl}                 & 56.7 & 74.9      & 54.2      & -       & 62.7   & -         & -    & -         \\
VAPO-Thinker-7B \cite{tian2025more}                 & 60.2 & 75.6      & 53.3      & -       & 63.0   & -         & -    & -         \\ 
 Qwen3-VL-Instruct-8B \cite{qwen3vl}            & 60.2 & 74.2      & 58.1      & 85.1    & 68.5   & 92.0      & 82.3 & 64.1      \\ \midrule
\rowcolor{front-color} OneThinker-8B                    & \textbf{70.6} & \textbf{77.6}      & \textbf{64.3}      & \textbf{86.6}    & \textbf{70.6}   & \textbf{96.5}      & \textbf{85.2} & \textbf{67.8}      \\ \bottomrule
\end{tabular}

}
\end{table*}

\begin{table*}[]
\vspace{-0.05in}
\caption{Performance of different models on video question answering benchmarks. For Qwen3-VL-Instruct-8B, we report our reproduced results under the same setting.}
\label{videoqa}
    \resizebox{\linewidth}{!}{%
    \setlength{\tabcolsep}{0.4mm}
     \renewcommand\arraystretch{1.2}
\begin{tabular}{@{}ccccccccc@{}}
\toprule
\multirow{2}{*}{\textbf{Models}} & \multirow{2}{*}{\textbf{Frames}} & \textbf{} & \multicolumn{6}{c}{\textbf{Video QA}}                                                \\ \cmidrule(l){3-9} 
                                 &                                  & VideoMMMU\cite{hu2025video} & MMVU(mc)\cite{zhao2025mmvu} & VideoMME\cite{fu2025video} & VideoHolmes\cite{cheng2025video} & LongVideoBench\cite{wu2024longvideobench} & LongVideo-Reason\cite{chen2025scaling} & VideoMathQA\cite{rasheed2025videomathqa} \\ \midrule
\rowcolor{Gray} GPT-4o \cite{hurst2024gpt}                            &                                  & 61.2      & 75.4     & 71.9     & 42.0         & 66.7           & -                & 20.2        \\
\rowcolor{Gray} Gemini 2.5 Pro \cite{{comanici2025gemini}}                  & -                                & 83.6      & -        & 84.3     & 45.0         & -              & -                & -           \\
\rowcolor{Gray} Seed1.5-VL \cite{guo2025seed1}                      & -                                & 81.4      & -        & 77.9     & -            & 74.0           & -                & -           \\ \midrule
VideoLLaMA3-7B \cite{zhang2025videollama}                  & -                                & -         & -        & 66.2     & -            & 59.8           & -                & -           \\
InternVideo2.5-8B \cite{wang2025internvideo2}                & -                                & -         & -        & 65.1     & -            & 60.6           & -                & 25.2        \\
VideoChat-R1-7B \cite{li2025videochat}                 & -                                & 46.4      & -        & 60.0     & 33.0         & -              & 67.2             & 27.6        \\
LongVILA-R1-7B  \cite{chen2025scaling}                 & -                                & 51.0      & -        & 65.1     & -            & 58.0           & 72.0             & 23.6        \\
Video-R1-7B \cite{feng2025video}                     & -                                & 52.4      & 64.2     & 61.4     & 36.5         & -              & 68.1             & 21.4        \\
Qwen3-VL-Instruct-8B \cite{qwen3vl}            & 128                              & 63.3      & 65.6     & 64.0     & 40.9         & 61.5           & 71.5             & 24.3        \\ \midrule
\rowcolor{front-color} OneThinker-8B                    & 128                              & \textbf{66.2}      & \textbf{70.5}     & \textbf{66.5}     & \textbf{48.7}         & \textbf{61.7}           & \textbf{79.2}             & \textbf{35.0}        \\ \bottomrule
\end{tabular}
\vspace{-0.25in}
}
\end{table*}



\textbf{Image QA.} OneThinker-8B consistently achieves top-tier performance for image QA across a diverse set of tasks spanning general knowledge, mathematics, science, and multimodal reasoning.
Compared with strong open-source models such as Vision-R1-7B, VAPO-Thinker-7B, and Qwen3-VL-Instruct-8B, our model attains superior results on these benchmarks.
For example, OneThinker reaches 70.6\% on MMMU, 77.6\% on MathVista, 64.3\% on MathVerse, and 70.6\% on MMStar, consistently outperforming all prior open-source competitors.
These results demonstrate that our unified reasoning framework can effectively generalize to a wide range of complex image QA scenarios.

\textbf{Video QA.} In video QA, OneThinker-8B shows strong superiority over video-focused reasoning models.
Across benchmarks including VideoMMMU, MMVU(mc), VideoMME, VideoHolmes, LongVideoBench, LongVideo-Reason, and VideoMathQA, OneThinker consistently ranks among the top performers.
For instance, it achieves 66.2\% on VideoMMMU, 70.5\% on MMVU(mc), and 66.5\% on VideoMME, outperforming specialized video reasoning models such as VideoChat-R1-7B, VideoLLaMA3-7B, and InternVideo2.5-8B.
Most notably, OneThinker obtains 79.2\% on LongVideo-Reason, substantially surpassing Video-R1-7B (67.2\%) and Qwen3-VL-Instruct-8B (71.5\%).
On VideoMathQA, a challenging video reasoning benchmark, OneThinker also leads all open-source models with a score of 35.0\%.
These results collectively verify that the effectiness of our proposed framework.

\begin{table}
\vspace{-0.2in}
\caption{Performance of different models on caption benchmarks.}
\vspace{0.1in}
\label{caption}
\centering
\small 
\setlength{\tabcolsep}{2.5mm}
\renewcommand\arraystretch{1}

\begin{tabular}{@{}ccccc@{}}
\toprule
\multirow{2}{*}{\textbf{Models}} & \multirow{2}{*}{\textbf{Frames}} & \multicolumn{2}{c}{\textbf{Image Caption}} & \textbf{Video Caption} \\ \cmidrule(l){3-5} 
                                 &                                  & MMSci-Caption\cite{li2024mmsci}         & MMT-Caption\cite{ying2024mmt}        & VideoMMLU-Caption\cite{song2025video}      \\ \midrule
\rowcolor{Gray} GPT-4o \cite{hurst2024gpt}                            & -                                & 27.0                  & -                  & 53.9                   \\ \midrule
LLaVA-1.5-7B \cite{liu2024improved}                    & -                                & 11.8                  & -                  & 22.3                   \\
Qwen3-VL-Instruct-8B \cite{qwen3vl}            & 128                              & 15.1                  & 47.3               & 20.0                   \\ \midrule
\rowcolor{front-color} OneThinker-8B                    & 128                              & \textbf{25.7}                  & \textbf{57.9}               & \textbf{28.0}                   \\ \bottomrule
\end{tabular}
\end{table}

\begin{table*}[]
\caption{Performance on temporal grounding benchmarks.}
\label{temporal}
    \resizebox{\linewidth}{!}{%
    \setlength{\tabcolsep}{1mm}
     \renewcommand\arraystretch{1.2}
     \small
\begin{tabular}{@{}cccccccccccccc@{}}
\toprule
\multirow{2}{*}{\textbf{Models}} & \multirow{2}{*}{\textbf{Frame}} & \multicolumn{4}{c}{\textbf{Charades} \cite{gao2017tall}}  & \multicolumn{4}{c}{\textbf{ActivityNet} \cite{krishna2017dense}}  & \multicolumn{4}{c}{\textbf{ANet-RTL} \cite{huang2024lita}} \\ \cmidrule(l){3-14} 
                                 &                                 & R@0.3    & R@0.5   & R@0.7   & mIoU   & R@0.3    & R@0.5    & R@0.7    & mIoU    & R@0.3    & R@0.5   & R@0.7   & mIoU   \\ \midrule
VTimeLLM \cite{huang2024vtimellm}                        & -                               & 55.3     & 34.3    & 14.7    & 34.6   & 44.8     & 29.5     & 14.2     & 31.4    & -        & -       & -       & -      \\
TimeSuite \cite{zeng2024timesuite}                       & -                               & 69.9     & 48.7    & 24.0    & -      & -        & -        & -        & -       & -        & -       & -       & -      \\
VideoChat-R1 \cite{li2025videochat}                     & -                               & 83.1     & \textbf{72.8}    & \textbf{51.5}    & \textbf{61.3}   & 50.4     & 32.2     & 16.2     & 34.3    & -        & -       & -       & -      \\
Temporal-RLT \cite{li2025reinforcement}                    & -                               & 80.2     & 68.3    & 44.5    & 57.9   & 56.9     & 38.4     & 20.2     & 39.1    & 40.2     & 22.7    & 10.9    & 26.3   \\
Time-R1 \cite{wang2025time}                         & -                               & 78.1     & 60.8    & 35.3    & -      & 58.6     & 39.0     & 21.4     & -       & -        & -       & -       & -      \\ 
Qwen3-VL-Instruct-8B \cite{qwen3vl}             & 128                             & 58.0     & 33.5    & 13.1    & 36.7   & 39.9     & 26.1     & 15.3     & 29.1    & 36.2     & 27.5    & 18.3    & 26.6   \\ \midrule
\rowcolor{front-color} OneThinker-8B                    & 128                             & \textbf{83.5}     & 68.3    & 45.3    & 59.9   & \textbf{65.0}     & \textbf{43.6}     & \textbf{25.7}     & \textbf{45.9}    & \textbf{62.0}     & \textbf{42.3}    & \textbf{22.7}    & \textbf{43.2}   \\ \bottomrule
\end{tabular}

}
\end{table*}

\textbf{Image and Video Caption.} On caption benchmarks, OneThinker maintains competitive or superior performance on both image and video captioning.
For image captioning, it achieves 25.7 on MMSci-Caption and 57.9 on MMT-Caption, markedly outperforming Qwen3-VL-Instruct-8B (15.1 and 47.3 respectively) and significantly improving over LLaVA-1.5-7B.
In video captioning, OneThinker reaches 28.0 on VideoMMLU-Caption, demonstrating effective video caption ability.
This unified captioning ability reflects the model’s strong visual descriptive skills.

\begin{table}
\vspace{-0.1in}
\caption{Performance on spatial grounding benchmarks.}
\vspace{0.1in}
\label{spatial}
    \setlength{\tabcolsep}{3.5mm}
     \renewcommand\arraystretch{1.1}
\small
\centering
\begin{tabular}{@{}ccccccccc@{}}
\toprule
\multirow{2}{*}{\textbf{Models}} & \multicolumn{3}{c}{\textbf{RefCOCO}\cite{kazemzadeh2014referitgame}} & \multicolumn{3}{c}{\textbf{RefCOCO+}\cite{kazemzadeh2014referitgame}} & \multicolumn{2}{c}{\textbf{RefCOCOg}\cite{yu2016modeling}} \\ \cmidrule(l){2-9} 
                                 & testA       & testB      & val       & testA       & testB       & val       & test              & val               \\ \midrule
Perception-R1 \cite{yu2025perception}                   & 91.4        & 84.5       & 89.1      & 86.8        & 74.3        & 81.7      & 85.4              & 85.7              \\
VLM-R1 \cite{shen2025vlm}                           & -           & -          & 90.5      & -           & -           & 84.3      & -                 & 87.1              \\
DeepEyes \cite{zheng2025deepeyes}                        & -           & -          & 89.8      & -           & -           & 83.6      & -                 & 86.7              \\
Qwen3-VL-Instruct-8B \cite{qwen3vl}            & 92.2        & 85.3       & 89.9      & 89.6        & 77.8        & 84.5      & 86.7              & 86.8              \\ \midrule
\rowcolor{front-color} OneThinker-8B                    & \textbf{93.7}        & \textbf{88.9}       & \textbf{92.0}      & \textbf{91.4}        & \textbf{82.7}        & \textbf{87.0}      & \textbf{88.8}              & \textbf{89.2}              \\ \bottomrule
\end{tabular}
\vspace{-0.1in}
\end{table}

\textbf{Temporal Grounding.} On temporal grounding tasks, OneThinker generally shows substantial improvements over existing temporal localization models.
For example, on Charades, OneThinker-8B achieves performance that is comparable to or better than previous models.
On ActivityNet, our model attains 65.0 R@0.3, 43.6 R@0.5, and 25.7 R@0.7, confirming superior temporal grounding abilities.
Furthermore, on the ANet-RTL benchmark, OneThinker achieves the best mIoU (43.2) among listed models.
These results demonstrate the model’s robust ability to reason about when events happen and accurately understand the fine-grained temporal information.

\textbf{Spatial Grounding.} For spatial grounding, OneThinker-8B also demonstrates state-of-the-art localization ability across the widely-used RefCOCO, RefCOCO+, and RefCOCOg benchmarks.
In RefCOCO testA/testB/val sets, it achieves 93.7 / 88.9 / 92.0, outperforming prior strong models such as Perception-R1, DeepEyes, and Qwen3-VL-Instruct-8B.
On RefCOCO+, OneThinker again leads with 91.4 / 82.7 / 87.0, consistently surpassing prior baselines by a large margin.
On the more challenging RefCOCOg benchmark, the model achieves 88.8 / 89.2 (test/val), showing strong comprehension of long and descriptive referring expressions.
These results highlight the model’s strong spatial grounding abilities.

\begin{table}[]
\vspace{-0.1in}
\caption{Performance on spatial-temporal grounding benchmarks.}
\vspace{0.1in}
\label{spatial-temporal}
\small
\centering
    \setlength{\tabcolsep}{4.5mm}
     \renewcommand\arraystretch{1.1}

\begin{tabular}{@{}cccccc@{}}
\toprule
\multirow{2}{*}{\textbf{Models}} & \multirow{2}{*}{\textbf{Frame}} & \multicolumn{4}{c}{\textbf{STVG} \cite{li2025llava}} \\ \cmidrule(l){3-6} 
                                 &                                 & tIoU@0.5 & tIoU & sIoU@0.5 & sIoU \\ \midrule
GroundingGPT \cite{li2024groundinggpt}                     & -                               & 7,1      & 12.2 & 2.9      & 9.2  \\
VTimeLLM \cite{huang2024vtimellm}                        & -                               & 7.1      & 15.5 & -        & -    \\
Grounded-VideoLLM \cite{wang2024grounded}               & -                               & 30.0     & 33.0 & -        & -    \\
Qwen3-VL-Instruct-8B \cite{qwen3vl}            & 128                             & 24.4     & 25.4 & 11.6     & 13.6 \\ \midrule
\rowcolor{front-color} OneThinker-8B              & 128                             & \textbf{34.9}     & \textbf{39.5} & \textbf{40.3}     & \textbf{36.7} \\ \bottomrule
\end{tabular}
\end{table}

\begin{table}[]
\caption{Performance on tracking benchmarks.}
\vspace{0.1in}
\centering
\label{tracking}
\small
    \setlength{\tabcolsep}{6mm}
     \renewcommand\arraystretch{1.1}
\begin{tabular}{@{}ccccc@{}}
\toprule
\multirow{2}{*}{\textbf{Models}} & \multicolumn{4}{c}{\textbf{GOT-10k} \cite{huang2019got}} \\ \cmidrule(l){2-5} 
                                 & AO     & R@0.3   & R@0.5   & R@0.7   \\ \midrule
R1-Track \cite{wang2025r1}                        & 68.0   & -       & 76.6    & -       \\
VideoChat-R1 \cite{li2025videochat}                    & 42.5   & -       & 30.6    & 3.9     \\
Qwen3-VL-Instruct-8B \cite{qwen3vl}            & 33.7   & 51.1    & 28.9    & 10.6    \\ \midrule
\rowcolor{front-color} OneThinker-8B                    & \textbf{73.0}   & \textbf{93.9}    & \textbf{84.4}    & \textbf{68.8}    \\ \bottomrule
\end{tabular}

\end{table}

\textbf{Spatial-Temporal Grounding.} On spatial-temporal grounding tasks, which require simultaneous localization in both space and time, OneThinker delivers substantial improvements over previous systems.
On the STVG benchmark, it achieves 34.9 tIoU@0.5, 39.5 tIoU, 40.3 sIoU@0.5, and 36.7 sIoU, outperforming Grounded-VideoLLM and Qwen3-VL-Instruct-8B by a large margin.
Such gains emphasize OneThinker’s capability to jointly reason about where and when events occur, even in complex videos involving multiple objects and temporal transitions.

\textbf{Tracking.} As for tracking tasks, OneThinker reaches a high 73.0 AO, 93.9 R@0.3, 84.4 R@0.5, and 68.8 R@0.7 on GOT-10k, outperforming previous models like R1-Track and VideoChat-R1.
Notably, our evaluation uses 32 frames for prediction, which is substantially more challenging than the 8-frame setting adopted by prior work VideoChat-R1.
This large improvement illustrates that the unified reasoning architecture also yields strong single-object tracking capabilities, enabling reliable localization over long temporal sequences.

\textbf{Image and Video Segmentation.} OneThinker achieves the highest mean performance across both image and video segmentation benchmarks.
For image segmentation (RefCOCO / RefCOCO+ / RefCOCOg), it obtains 75.8 / 67.1 / 70.8 cIoU on the val set, significantly outperforming PixelLM-7B, LISA-7B, VISA-13B, Seg-R1-7B.
For video segmentation, it reaches 48.8 J, 56.7 F, and 52.7 J\&F on MeViS, surpassing all previous open-source models.
On ReasonVOS, it delivers 51.1 J, 58.7 F, and 54.9 J\&F, again achieving the best performance across all competitors.
These results demonstrate the model’s fine-grained visual understanding ability in both static and dynamic environments.

Overall, OneThinker serves as a unified multimodal reasoning generalist that achieves strong
performance across all major visual understanding tasks, showing strong potential for scalable and
generalizable visual reasoning.

\begin{table*}[]
\caption{Performance on segmentation benchmarks. For RefCOCO series, we report results on the val set.}
\label{segmentation}
    \resizebox{\linewidth}{!}{%
    \setlength{\tabcolsep}{1.1mm}
     \renewcommand\arraystretch{1.2}
\small
\begin{tabular}{@{}cccccccccc@{}}
\toprule
\multirow{3}{*}{\textbf{Models}} & \multicolumn{3}{c}{\textbf{Image Segmentation}} & \multicolumn{6}{c}{\textbf{Video Segmentation}}           \\ \cmidrule(l){2-10} 
                                 & RefCOCO \cite{kazemzadeh2014referitgame}       & RefCOCO+ \cite{kazemzadeh2014referitgame}      & RefCOCOg \cite{yu2016modeling}      & \multicolumn{3}{c}{MeViS \cite{ding2023mevis}} & \multicolumn{3}{c}{ReasonVOS \cite{bai2024one}} \\ \cmidrule(l){2-10} 
                                 & cIoU          & cIoU           & cIoU           & J       & F       & J\&F  & J        & F        & J\&F    \\ \midrule
PixelLM-7B \cite{ren2024pixellm}                       & 73.0          & 66.3           & 69.3           & -       & -       & -     & -        & -        & -       \\
LISA-7B \cite{lai2024lisa}                          & 74.1          & 62.4           & 66.4           & -       & -       & -     & 29.1     & 33.1     & 31.1    \\
VISA-13B \cite{yan2024visa}                        & 72.4          & 59.8           & 65.5           & -       & -       & 44.5  & -        & -        & -       \\
Seg-R1-7B \cite{you2025seg}                       & 74.3          & 62.6           & \textbf{71.0}          & -       & -       & -     & -        & -        & -       \\
ReferFormer \cite{wu2022language}                      & -             & -              & -              & 29.8    & 32.2    & 31.0  & 30.2     & 35.6     & 32.9    \\
VideoLISA-3.8B \cite{bai2024one}                   & -             & -              & -              & 41.3    & 47.6    & 44.4  & 45.1     & 49.9     & 47.5    \\
Qwen3-VL-Instruct-8B \cite{qwen3vl} + SAM2 \cite{ravi2024sam}      & 73.2          & 66.2           & 68.3           & 19.4    & 26.4    & 22.9  & 16.6     & 22.7     & 19.6    \\ \midrule
\rowcolor{front-color} OneThinker-8B                    & \textbf{75.8}          & \textbf{67.1}           & 70.8           & \textbf{48.8}    & \textbf{56.7}    & \textbf{52.7}  & \textbf{51.1}     & \textbf{58.7}     & \textbf{54.9}   \\ \bottomrule
\end{tabular}

}
\vspace{-0.1in}
\end{table*}

\subsection{Ablation Study}

In this section, we design three variants of OneThinker to verify the effectiveness of different components in our framework: 
(1) OneThinker-8B-SFT, which is trained only with SFT without
RL;  
(2) OneThinker-8B-GRPO, which replaces our proposed EMA-GRPO with the original GRPO
algorithm;  
(3) OneThinker-8B-DrGRPO, which adopts the Dr.GRPO \cite{liu2025understanding} algorithm for RL training.

As shown in \cref{ablation}, all ablated variants perform worse than OneThinker-8B across
all tasks. Compared with the SFT baseline, RL consistently improves performance,
demonstrating its effectiveness across diverse visual tasks. Replacing EMA-GRPO with standard GRPO or DrGRPO results in noticeable degradation,
demonstrating the importance of addressing the intra-task imbalance and inter-task imbalance issues.  
Overall, This ablation study confirms the effectiveness of our unified RL framework and the proposed EMA-GRPO algorithm.

\begin{table*}[]
\vspace{-0.2in}
\caption{Ablation study. Results are averaged over benchmarks within each task for comparison.
For tasks with multiple evaluation metrics, we report the following: mIoU for temporal grounding, the mean of tIoU and sIoU for spatio-temporal grounding, AO for tracking, and the mean of cIoU (image) and J\&F (video) for segmentation.}

\label{ablation}
    \resizebox{\linewidth}{!}{%
    \setlength{\tabcolsep}{3mm}
     \renewcommand\arraystretch{1.2}
\tiny

\begin{tabular}{@{}ccccccc@{}}
\toprule
\textbf{Models}      & \textbf{QA}  & \textbf{\begin{tabular}[c]{@{}c@{}}Temporal\\  Grounding\end{tabular}} & \textbf{\begin{tabular}[c]{@{}c@{}}Spatial \\ Grounding\end{tabular}} & \textbf{\begin{tabular}[c]{@{}c@{}}S.-T. \\ Grounding\end{tabular}} & \textbf{Tracking} & \textbf{Segmentation} \\ \midrule
Qwen3-VL-Instruct-8B & 65.0                   & 30.8                                                                   & 86.6                                                                  & 19.5                                                                & 33.7              & 50.0                  \\
OneThinker-8B-SFT    & 67.0                & 31.8                                                                   & 87.8                                                                  & 27.1                                                                & 48.1              & 62.8                  \\
OneThinker-8B-GRPO   & 67.2                 & 46.9                                                                   & 86.5                                                                  & 34.5                                                                & 65.5              & 62.3                  \\
OneThinker-8B-DrGRPO & 67.6                   & 46.3                                                                   & 88.2                                                                  & 34.0                                                                & 67.8              & 61.2                  \\ \midrule
\rowcolor{front-color} OneThinker-8B        & 69.8            & 49.7                                                                   & 89.2                                                                  & 38.1                                                                & 73.0              & 64.2                  \\ \bottomrule
\end{tabular}
}
\vspace{-0.1in}
\end{table*}

\subsection{Benefits of Unified Training Analysis}

To further investigate the potential benefits and knowledge sharing of cross-task and cross-modal learning,
we conduct an analysis by selectively removing data from specific task categories during training.  
We design three variants:  
(1) OneThinker-wo-spatial-grounding, which excludes spatial grounding data;  
(2) OneThinker-wo-temporal-grounding, which removes all temporal grounding data;  
(3) OneThinker-wo-ImageQA, which omits all image QA samples.

As shown in \cref{benefit}, removing either spatial or temporal grounding
leads to a noticeable drop in performance across other tasks.
In particular, the absence of temporal grounding significantly degrades results on video QA
and tracking, indicating that temporal grounding may enhance the model’s temporal perception
and sequential reasoning ability.  
Similarly, removing spatial grounding results in lower accuracy on both image QA and segmentation,
indicating that spatial localization tasks may contribute valuable structural and positional cues that
benefit broader visual reasoning.

Moreover, excluding ImageQA causes the severe performance drop on video QA.
We attribute this to the generally higher quality and greater diversity of image QA datasets,
which help the model develop stronger general reasoning and recognition capabilities that
transfer well to video understanding.  
This observation confirms that knowledge learned from static images can generalize to dynamic
video scenarios, reflecting the benefit of cross-modal transfer.

Overall, these results suggest that certain tasks and modalities can benefit from others during joint training in OneThinker.
By jointly training on diverse visual tasks,
OneThinker effectively shares knowledge across domains and emerges as a multimodal reasoning
generalist.

\begin{table}[]
\caption{Benefits between tasks and modalities analysis.}
\vspace{0.1in}
\label{benefit}
\centering
    \setlength{\tabcolsep}{5mm}
     \renewcommand\arraystretch{1.3}

\small
\begin{tabular}{@{}ccccc@{}}
\toprule
\textbf{Variants}      & \textbf{Image QA} & \textbf{Video QA} & \textbf{Tracking} & \textbf{Segmentation} \\ \midrule
OneThinker-wo-spatial-grounding  & 76.6             & 60.3             & 71.0              & 62.9                  \\
OneThinker-wo-temporal-grounding & 77.2             & 59.5             & 67.2              & 63.3                  \\ \midrule
OneThinker-wo-ImageQA & -                & 58.2             & 72.3              & 63.9                  \\ \midrule
\rowcolor{front-color} OneThinker             & 77.4             & 61.1             & 73.0              & 64.2                  \\ \bottomrule
\end{tabular}

\end{table}

\subsection{Zero-shot Generalization to Unseen Tasks}

\begin{figure}[h]
  \centering
  \includegraphics[width=0.7\linewidth]{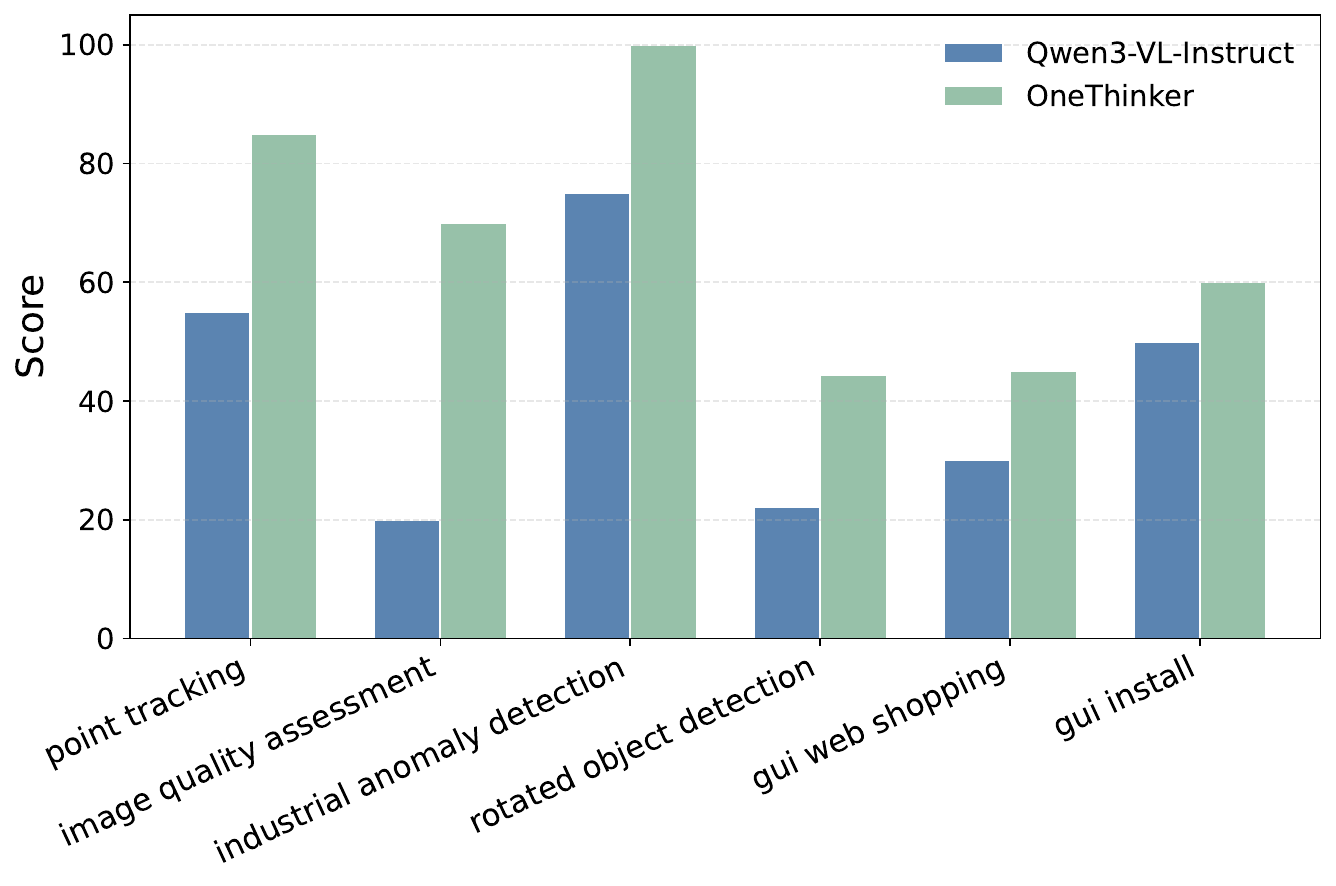}
  \vspace{-0.1in}
  \caption{
    Performance on unseen visual tasks.
  }
  \label{zeroshot}
  \vspace{-0.2in}
\end{figure}

We further evaluate OneThinker’s zero-shot generalization on unseen visual understanding tasks. These unseen tasks are selected from MMT-Bench \cite{ying2024mmt}, which contains 162 diverse visual tasks.
As shown in \cref{zeroshot}, OneThinker-8B clearly outperforms Qwen3-VL-Instruct-8B
across multiple unseen tasks, such as point tracking, image quality assessment, GUI tasks and rotated obejct detection.
These results demonstrate that unified multimodal reasoning enables the model to generalize beyond
its training tasks, showing promising transferability to novel real-world scenarios.

\section{Conclusion}
\label{sec:conclusion}

In this work, we present OneThinker, an all-in-one multimodal reasoning model that unifies diverse visual foundation tasks for images and videos.  
To support training, we construct OneThinker-600k dataset for RL training and its CoT-annotated subset OneThinker-SFT-340k for SFT cold start.  
We further propose EMA-GRPO, an RL algorithm that balances optimization across heterogeneous visual tasks through task-wise adaptive reward normalization.  
Extensive experiments demonstrate that OneThinker achieves strong performance across tasks.  
We hope this work takes a step toward scalable and unified multimodal reasoning generalist.

{
    \small
    \bibliographystyle{unsrt}
    \bibliography{iclr2026_conference}
}

\clearpage

\appendix


\section{Reasoning Examples}

\begin{figure*}[h]
  \centering
  \includegraphics[width=0.9\linewidth]{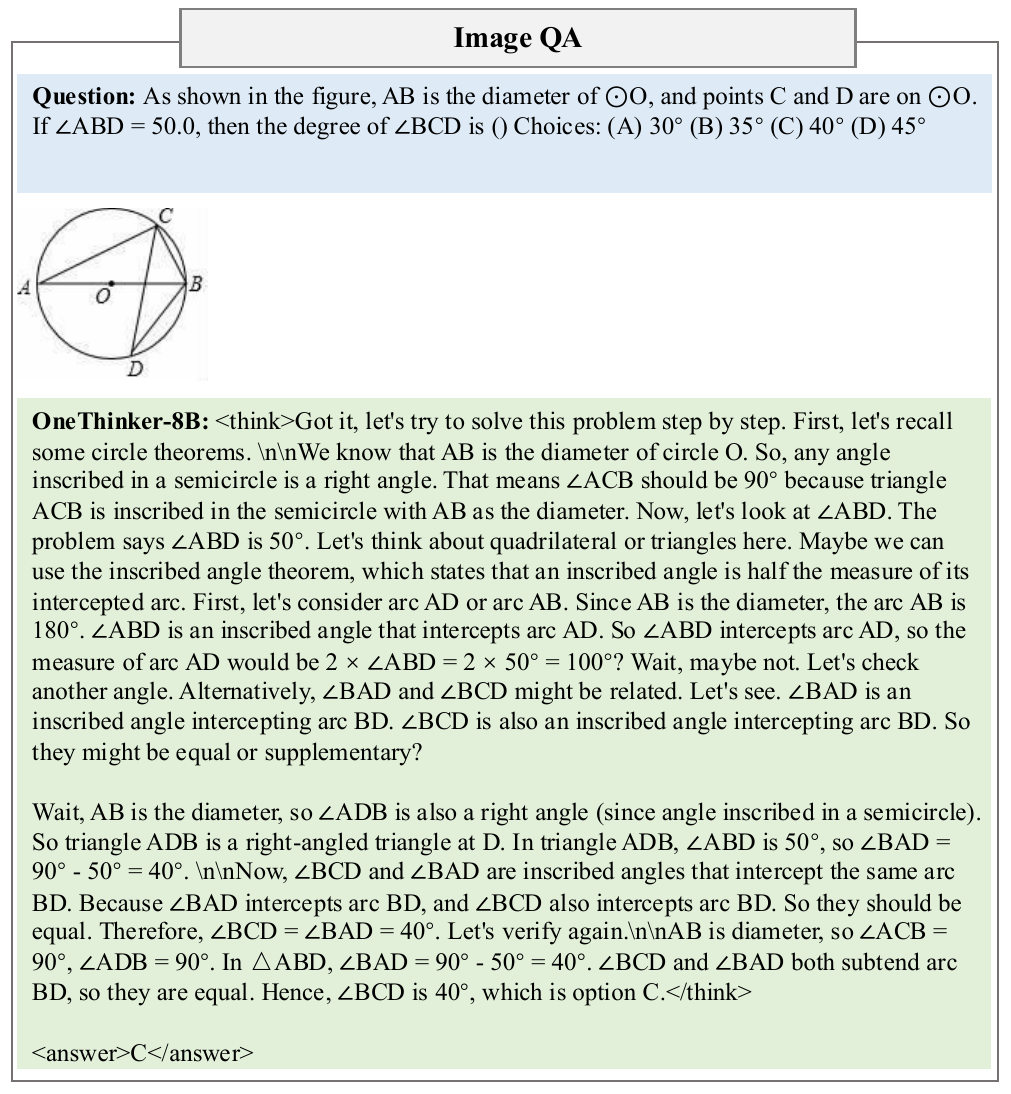}
  \caption{
Reasoning example of image question answering task.
  }
\end{figure*}

\begin{figure*}[h]
  \centering
  \includegraphics[width=0.95\linewidth]{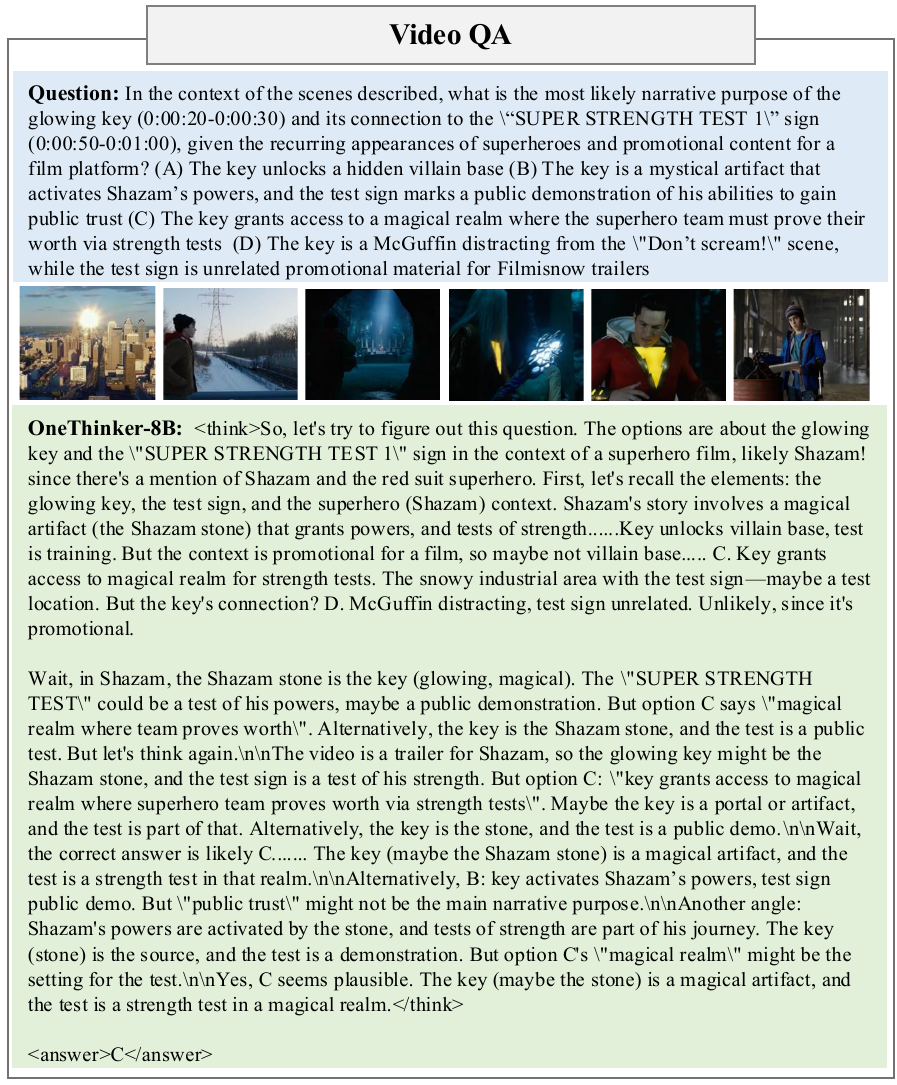}
  \caption{
Reasoning example of video question answering task.
  }
\end{figure*}

\begin{figure*}[h]
  \centering
  \includegraphics[width=0.95\linewidth]{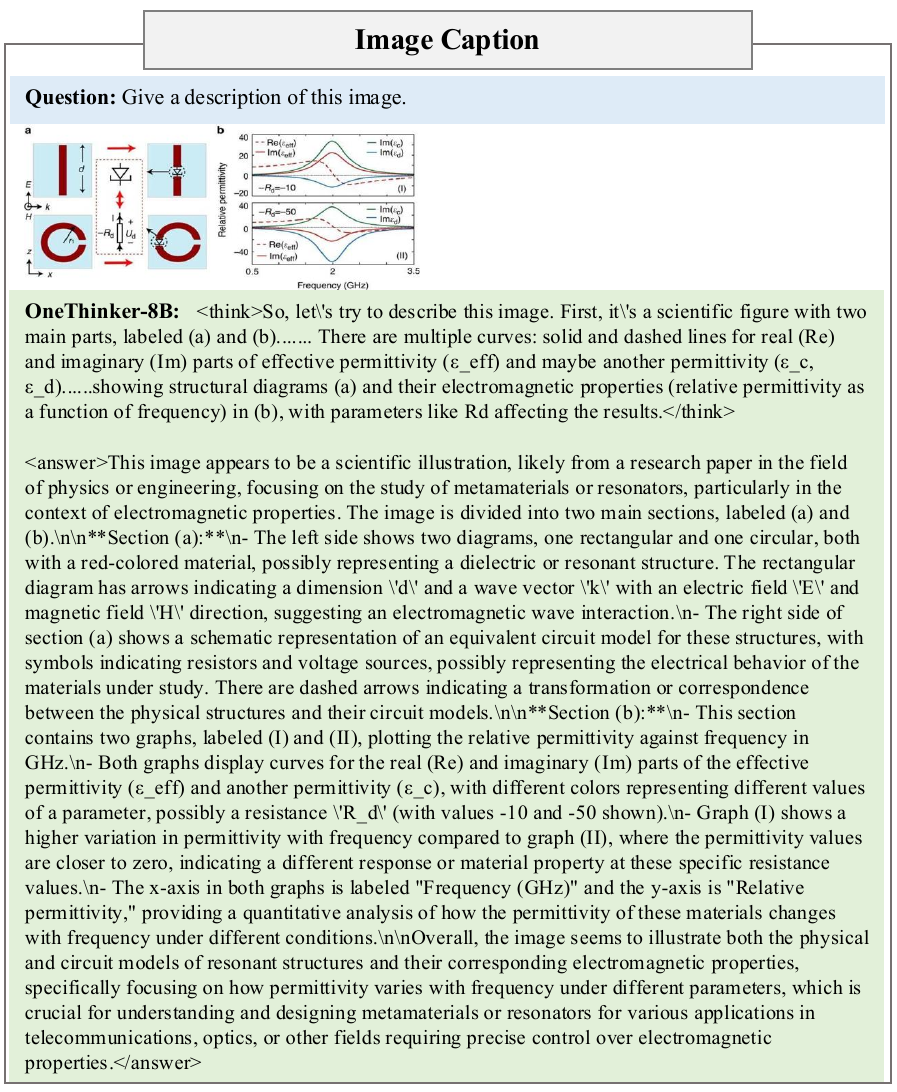}
  \caption{
Reasoning example of image caption task.
  }
\end{figure*}

\begin{figure*}[h]
  \centering
  \includegraphics[width=0.95\linewidth]{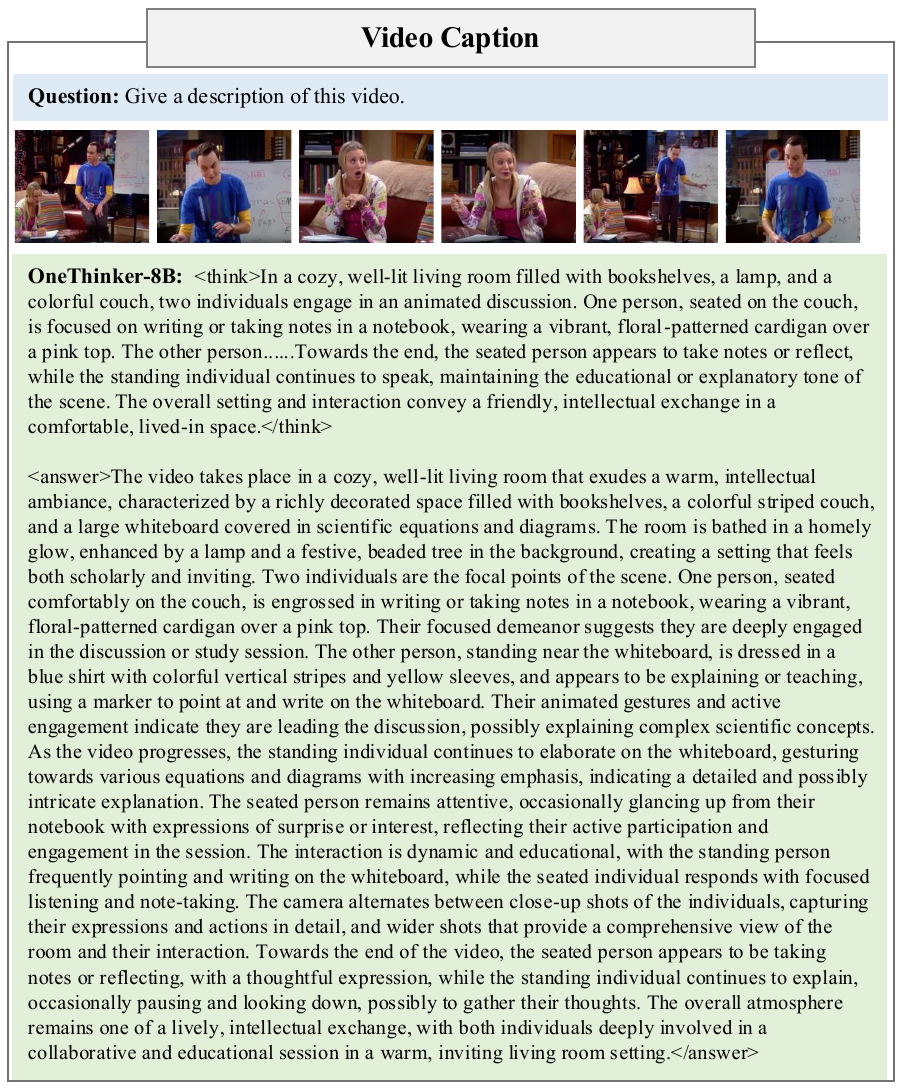}
  \caption{
Reasoning example of video caption task.
  }
\end{figure*}

\begin{figure*}[h]
  \centering
  \includegraphics[width=0.95\linewidth]{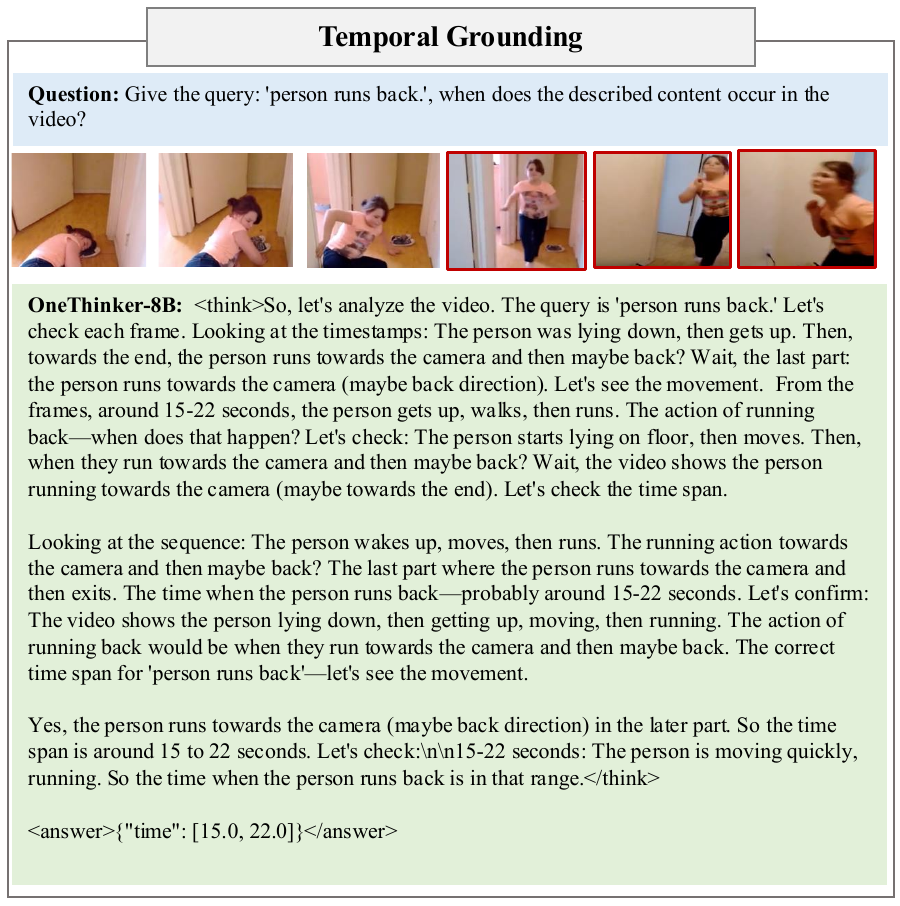}
  \caption{
Reasoning example of temporal grounding task.
  }
\end{figure*}

\begin{figure*}[h]
  \centering
  \includegraphics[width=0.95\linewidth]{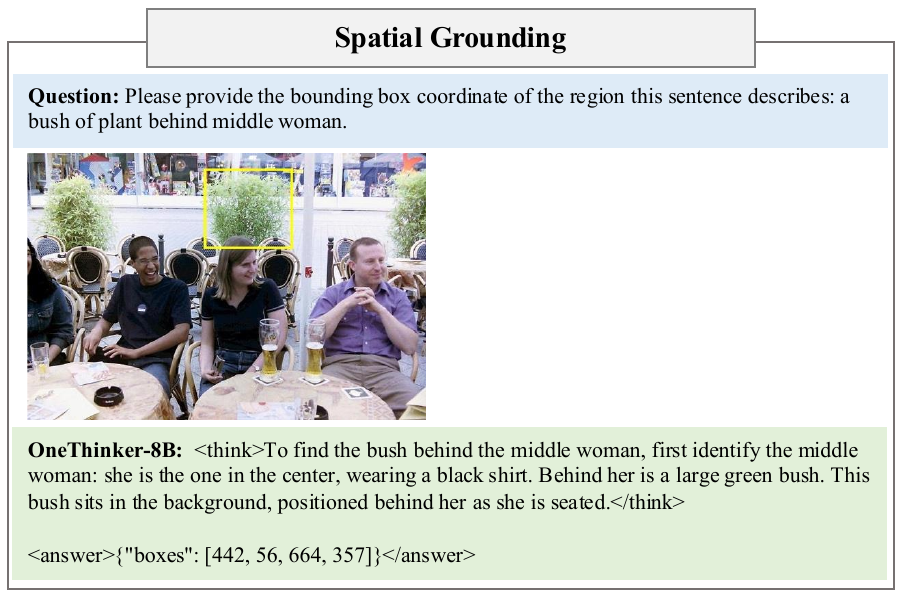}
  \caption{
Reasoning example of spatial grounding task.
  }
\end{figure*}

\begin{figure*}[h]
  \centering
  \includegraphics[width=0.95\linewidth]{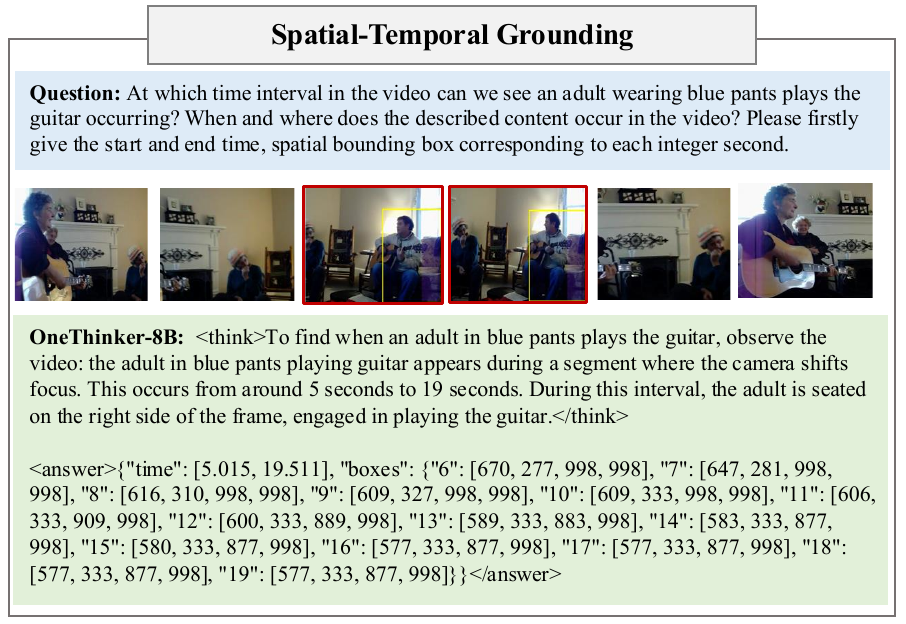}
  \caption{
Reasoning example of spatial-temporal grounding task.
  }
\end{figure*}

\begin{figure*}[h]
  \centering
  \includegraphics[width=0.95\linewidth]{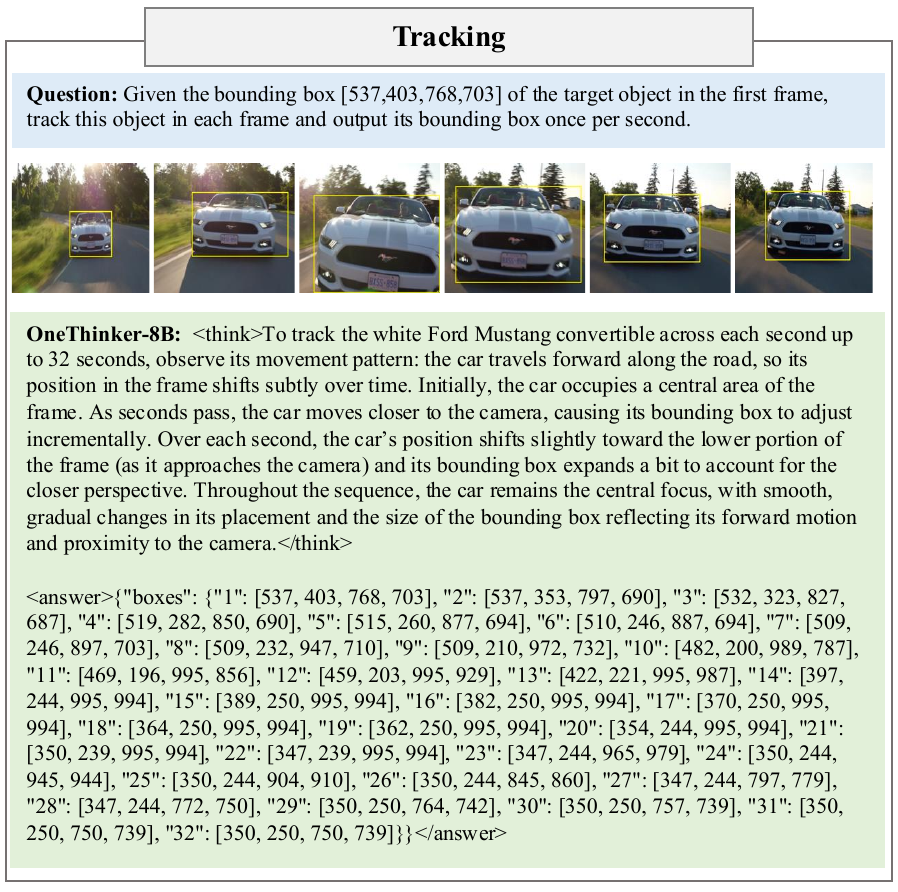}
  \caption{
Reasoning example of tracking task.
  }
\end{figure*}

\begin{figure*}[h]
  \centering
  \includegraphics[width=0.95\linewidth]{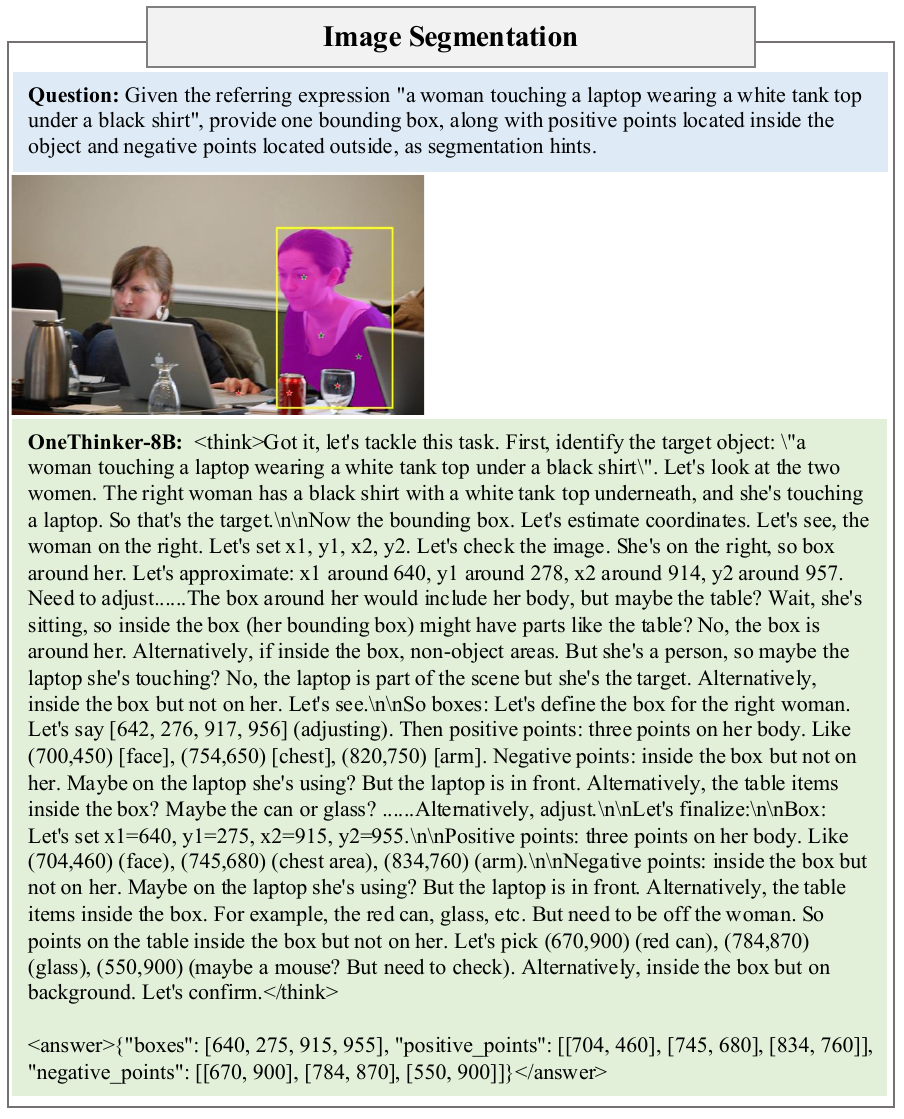}
  \caption{
Reasoning example for an image segmentation task. The resulting answer will be forwarded to SAM2 to produce the mask.
  }
\end{figure*}

\begin{figure*}[h]
  \centering
  \includegraphics[width=0.95\linewidth]{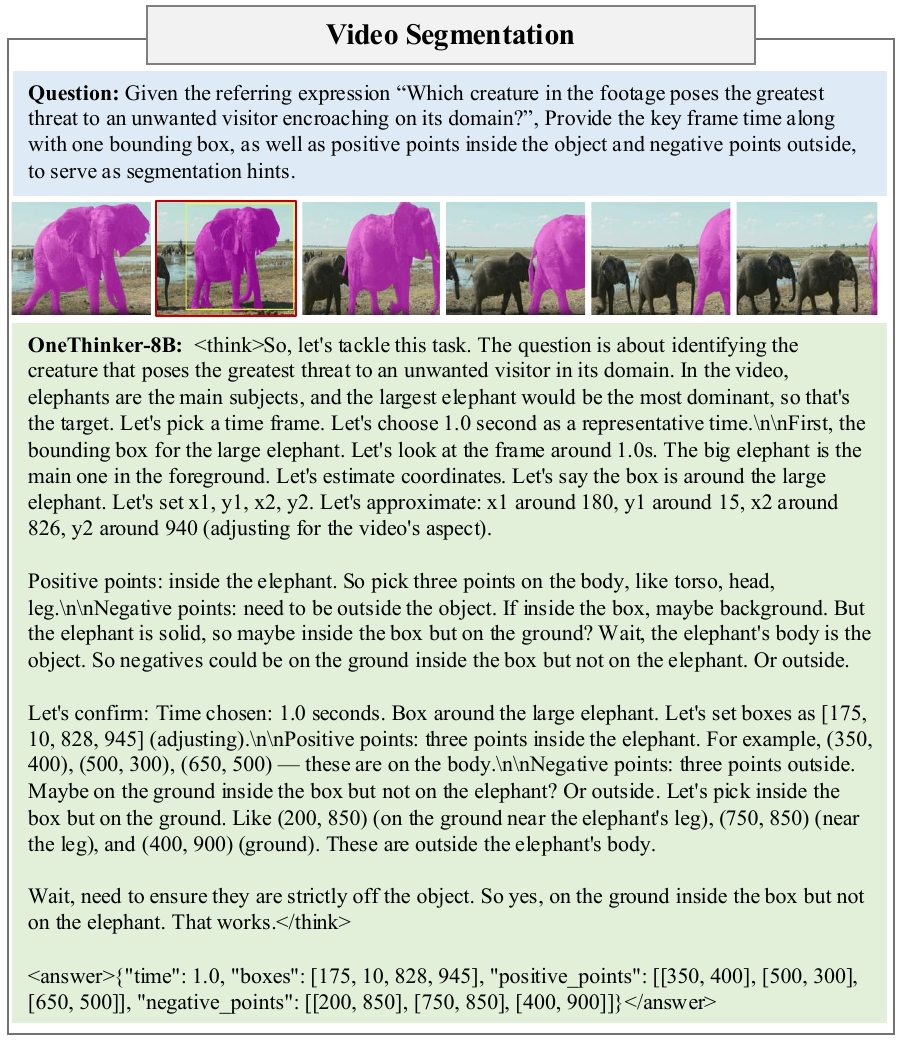}
  \caption{
Reasoning example for an video segmentation task. The resulting answer will be forwarded to SAM2 to produce the mask.
  }
\end{figure*}

\clearpage

\section{Prompt Template}

\begin{figure*}[h]
  \centering
  \includegraphics[width=0.9\linewidth]{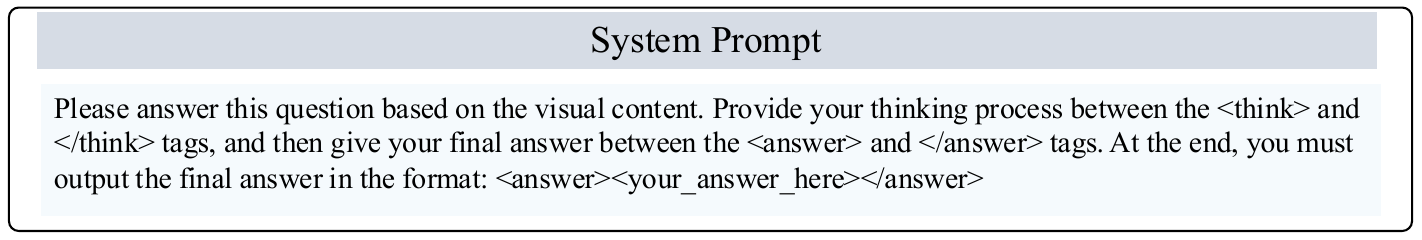}
  \caption{
System prompt for all tasks.
  }
\end{figure*}

\vspace{0.5in}

\begin{figure*}[h]
  \centering
  \includegraphics[width=0.67\linewidth]{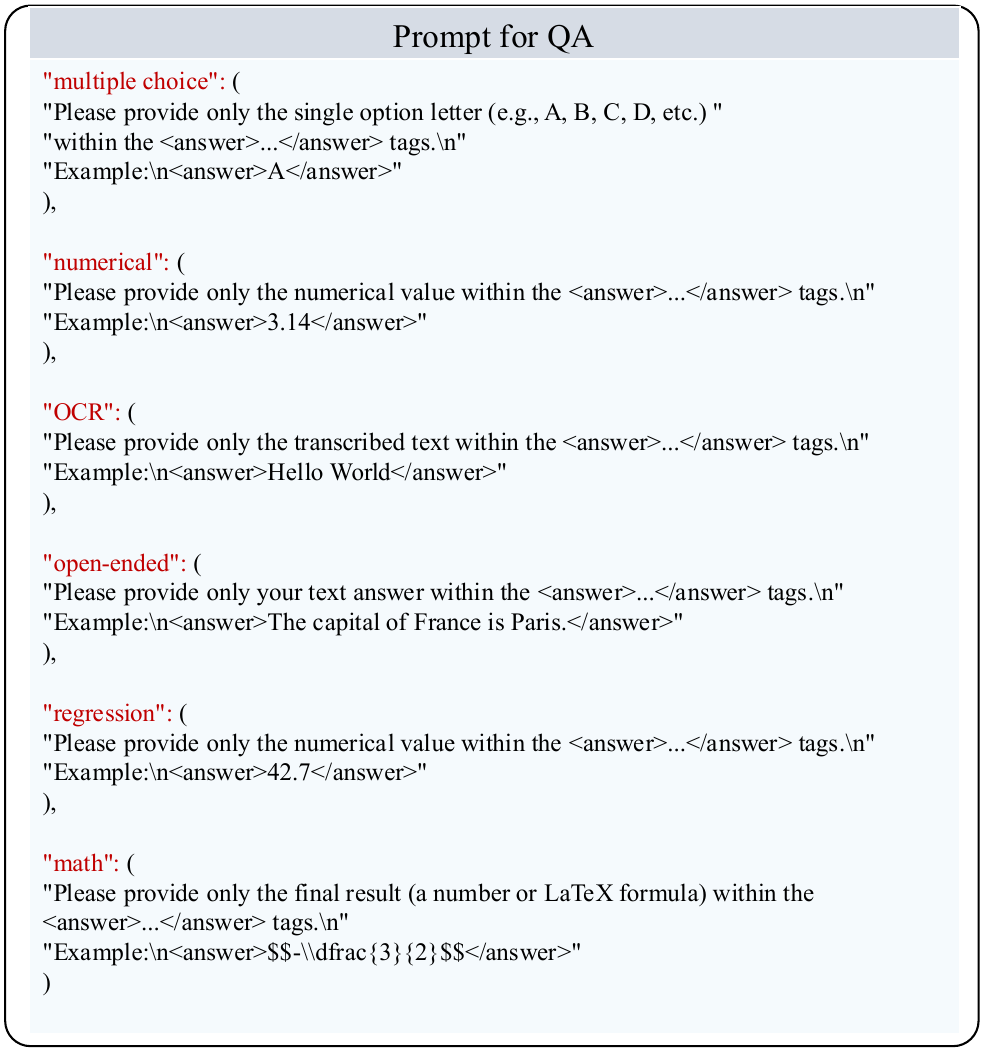}
  \caption{
Prompt for QA tasks.
  }
\end{figure*}

\begin{figure*}[h]
  \centering
  \includegraphics[width=0.67\linewidth]{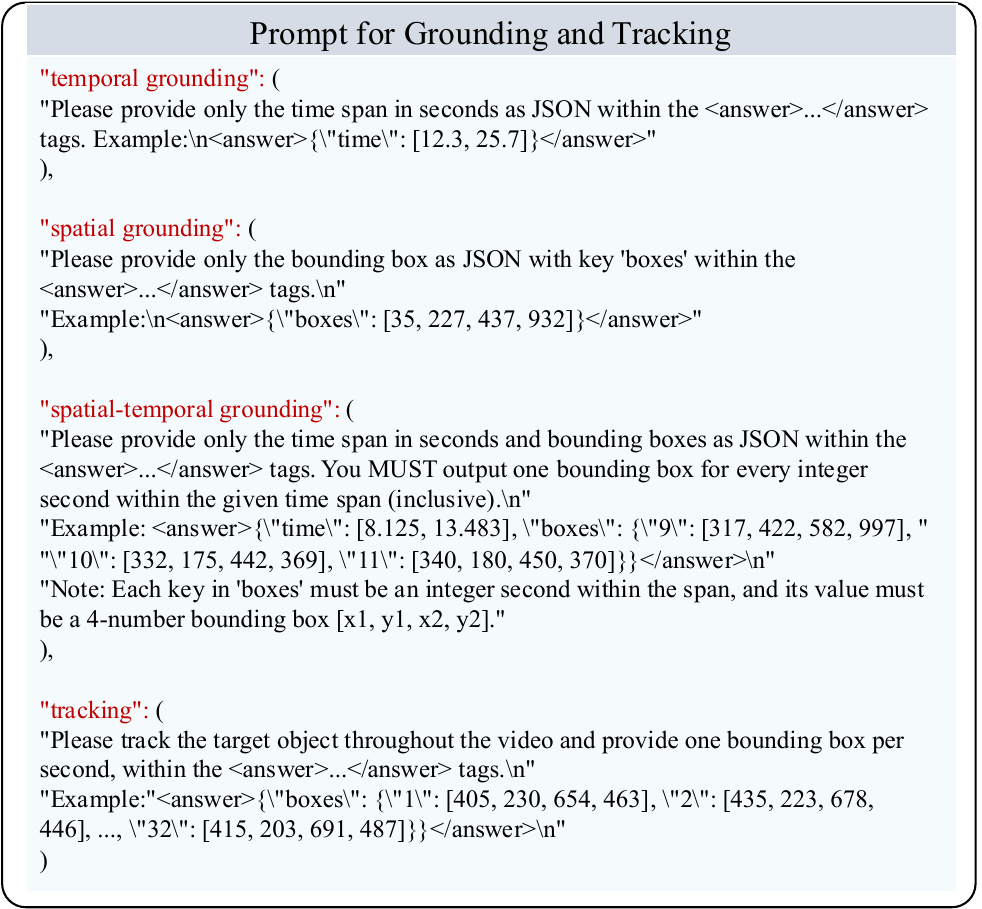}
  \caption{
Prompt for grounding and tracking tasks.
  }
\end{figure*}

\vspace{-0.5in}

\begin{figure*}[h]
  \centering
  \includegraphics[width=0.67\linewidth]{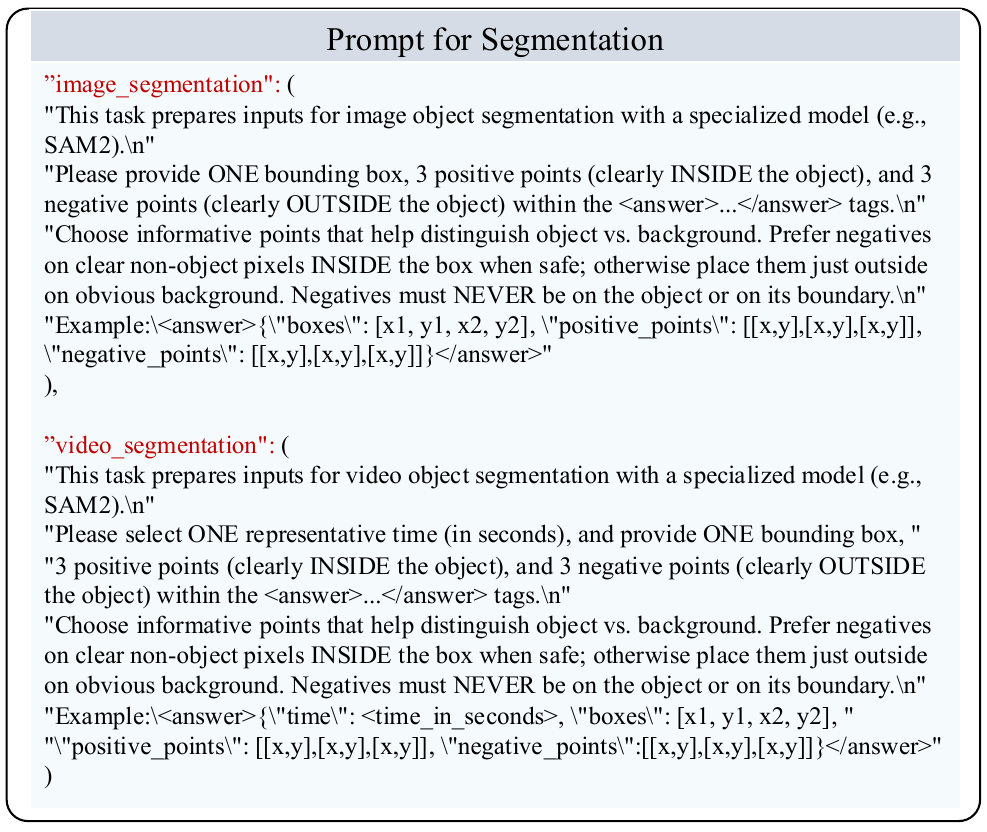}
  \caption{
Prompt for segmentation tasks.
  }
\end{figure*}

\end{document}